\documentclass{ws-rv975x65}
\pdfoutput=1
\usepackage{subfigure}     
\usepackage{ws-rv-thm}     
\usepackage{ws-rv-van}     
\usepackage{ wasysym }  
\makeindex

\graphicspath{{Figures}}  
\usepackage{epsfig}  
\usepackage{psfrag}  
\usepackage{float}
\usepackage{color}  

\usepackage[textsize=footnotesize, bordercolor=white,backgroundcolor=gray!30,linecolor=black,colorinlistoftodos]{todonotes}  

\usepackage[normalem]{ulem} 
\usepackage{multirow}  
\usepackage{array}
\usepackage{hyperref}

\begin{document}
\chapter[Snake-Like Robots for MIS]{Snake-Like Robots for \\Minimally Invasive, Single Port, and Intraluminal Surgeries} \label{mis_snakes_chapter}
\author[A. Orekhov, C. Abah, N. Simaan]{Andrew L. Orekhov, Colette Abah, Nabil Simaan}
\address{Vanderbilt University, Department of Mechanical Engineering,\\
405 Olin Hall, PMB 351592, Nashville, TN 37235, \\
{\{andrew.orekhov, c.abah, nabil.simaan\}@vanderbilt.edu}}
\begin{abstract}
The surgical paradigm of Minimally Invasive Surgery (MIS) has been a key driver to the adoption of robotic surgical assistance. Progress in the last three decades has led to a gradual transition from manual laparoscopic surgery with rigid instruments to robot-assisted surgery. In the last decade, the increasing demand for new surgical paradigms to enable access into the anatomy without skin incision (intraluminal surgery) or with a single skin incision (\textit{Single Port Access surgery - SPA}) has led researchers to investigate snake-like flexible surgical devices. In this chapter, we first present an overview of the background, motivation, and taxonomy of MIS and its newer derivatives. Challenges of MIS and its newer derivatives (SPA and intraluminal surgery)  are outlined along with the architectures of new snake-like robots meeting these challenges. We also examine the commercial and research surgical platforms developed over the years, to address the specific functional requirements and constraints imposed by operations in confined spaces. The chapter concludes with an evaluation of open problems in surgical robotics for intraluminal and SPA, and a look at future trends in surgical robot design that could potentially address these unmet needs.
\end{abstract} 
\body
\section{Minimally Invasive Surgery: History and Drivers}\label{mis_history}
\par Minimally invasive surgery (MIS) is a surgical paradigm that evolved slowly over the last two centuries. This slow, yet steady progression depended at every step on engineering innovation to support surgical innovation. First came minimally invasive visualization tools starting with early works of Bozzini's cystoscope\cite{scott1969development} (1805) followed by the first endoscope by Desormeaux (1853) \cite{samplaski2009two}, and succeeded by Kelling's early attempts at laparoscopy \cite{litynski1997laparoscopy} (1901). The invention of the Hopkins Rod endoscope and subsequently the availability of digital cameras have allowed surgeons to obtain better view of the anatomy while removing the need to peer through the endoscope lens; thus, freeing surgeons to operate more ergonomically and allowing them to use their hands to manipulate surgical instruments. More importantly, the use of the camera and monitor display allowed the surgeon and an assistant to observe the site of surgery in real-time, while collaborating on completing surgical tasks. This technological advancement has ushered in the age of minimally invasive laparoscopic surgery starting in the early 1980's with the first laparoscopic appendectomy by Kurt Semm \cite{semm1983endoscopic} and culminating in 1994 with the first laprascopic Whipples procedure \cite{gagner1994laparoscopic}.
\par The steady progression towards reduction of invasiveness has benefited patients by reducing blood loss, scarring, wound site infection, hernia, pain, and post-operative recovery. Unfortunately, these benefits to the patients came at a price to surgeons who have to contend with steep learning curve owing to the inverse kinematic mapping of hand-to-tooltip motion due to incisional constraints\footnote{Such constraints allow only 4 degrees-of-freedom (insertion along and rotation about the tool axis combined with two tilting motions about two perpendicular axes belonging to the local tangent plane of the skin at the incision point)}, lack of dexterity, and loss of sensory information compared to open surgery.
These challenges, combined with the need for the manipulation of multiple tools through multiple ports have led to the inception of robot-assisted multi-port surgery, which grew steadily starting the mid 1990's to the first release of the da-Vinci system by Intuitive Surgical in the early 2000's.
\par
\begin{figure}[htbp]
\center
\includegraphics[width=0.4\textwidth]{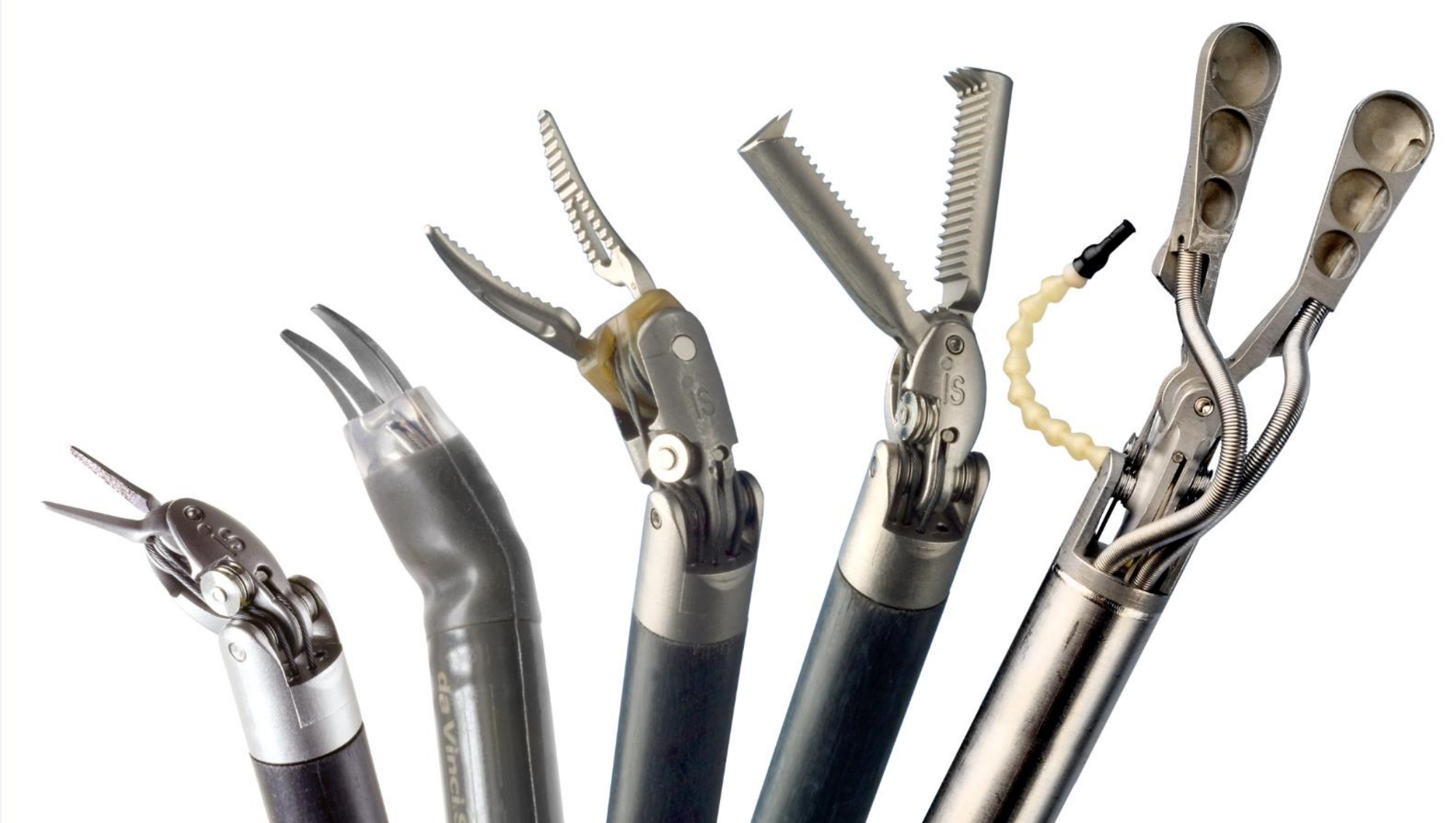}
\caption{An array of dexterous tools for MIS showing EndoWrist on the left. Image courtesy of Intuitive Surgical, Inc. \cite{daVinci_images} }
\label{fig:davinci_endowrist}
\end{figure}

 The success of robot-assisted MIS in reducing surgeon's cognitive and physiological loading are largely attributed to technological improvements in providing 3D stereo visualization and the introduction of dexterous wrists at the distal end of rigid tools. Examples of such distal dexterity wrists is shown in Figure \ref{fig:davinci_endowrist}. These distal dexterity wrists have allowed surgeons to achieve complex tissue manipulation and suturing that otherwise are very hard to achieve using manual laparoscopic tools.
\par Despite the immense progress made using rigid instruments with dexterous wrists, the continued demand to explore new surgical paradigms allowing access into the anatomy with no skin incisions (e.g. by using natural orifices) or by using a single incision have highlighted the limitations of such devices\cite{rattner2006asge, romanelli2009single, tiwari2010vivo, samarasekera2014robotic, morelli2015vinci, bae2016current}. To overcome these limitations, researchers in the past decade have explored the use of snake-like devices for surgery. The development of these devices have gone hand-in-hand with the exploration of new minimally invasive surgical paradigms. This chapter provides an overview of works in the areas of minimally invasive surgery with emphasis on MIS in confined spaces and its derivatives as highlighted in the next section.

\section{A Brief Taxonomy of MIS and its Derivatives}\label{sec:taxonomy_MIS_derivatives}
Multi-port MIS requires several small incisions that generally heal well, but can also be associated with pain, scarring and potential wound infection and/or hematoma. To ameliorate surgical outcomes, several new surgical approaches have been proposed to reduce or eliminate the number of surgical access incisions. Table~\ref{tab:surgery_taxonomy} follows the classification of Vitiello et al. \cite{Yang2013_notes_review} in presenting a taxonomy of MIS. The classification is based on the nature of the access route to the surgical site. \textit{Extraluminal} surgery involves the use of skin incision/s to access internal anatomy. This category of procedures is further categorized into multi-port and single port access (SPA) procedures. The vast majority of current MIS procedures are extraluminal multi-port procedures requiring 3-6 incisions. A small fraction of procedures currently can use the SPA approach and commercial robotic systems for SPA have only been released recently (with Intuitive Surgical's da-Vinci single port being the first FDA approved and clinically used\cite{morelli2015vinci}). During SPA a single incision is made and a multi-port trocar allowing the use of multiple tools (2 tools and a stereo camera and insufflation) is placed in the abdomen to provide surgical access.  Such single port typically is 30-40 mm in diameter and can be used at the umbilicus with hardly any visible scarring after the procedure.
\par The second major category is \textit{Intraluminal} surgery, which includes \textit{endoluminal} and \textit{transluminal} procedures. Endoluminal procedures involve operating within a bodily lumen that can be accessed using a natural orifice (therefore, such procedures are often referred to as natural orifice surgeries). Examples of endoluminal procedures include trans-oral MIS surgery of the airways \cite{Dowthwaite2012,Leoncini2014,Friedrich2017} esophageal surgery \cite{mueller2016endoluminal, sengupta2016advances,Stavropoulos2014,Inoue2015}, transanal colorectal microsurgery \cite{Tanaka2008,Arezzo2014}, transurethral prostate and bladder procedures \cite{richards2014importance,Mitchell2016}. Transluminal procedures use access along bodily lumens but require the creation of an incision in the lumen walls to access the surgical site. These procedures are typically referred to as natural orifice transluminal endoscopic surgeries (NOTES). Examples of such procedures include trans-gastric  and trans-vaginal abdominal surgery \cite{Palanivelu2008,Lehman2009,Choi2017}, trans-esophageal thoracic surgery\cite{Dake1994,Kato2001}, and trans-anal mesorectal surgery \cite{araujo2015transanal,figueredo2017robotic}.
\begin{table}[h]
\tbl{Taxonomy of Minimally Invasive Surgery. \vspace{0.5\baselineskip}}
{\begin{tabular}{cccc}
\hline
\multicolumn{2}{|c|}  {\textbf{Extra-luminal}} & \multicolumn{2}{c|}  {\textbf{Intra-luminal}} \\[2pt]  \cline{1-4}
\multicolumn{1}{|c|}{\textbf{Multi Port}} & \multicolumn{1} {c|}{\textbf{Single Port}} & \multicolumn{1}{c|}{\textbf{Endoluminal}} & \multicolumn{1}{c|}{\textbf{Transluminal}} \\[0pt]  \hline
\multicolumn{1}{|c|}{\parbox{2.4cm}{\vspace{1mm}Multiple ports (3-6) ports used to access anatomy.  One tool per access port.\vspace{1mm}}} &
\multicolumn{1}{c|}{\parbox{2.4cm}{A single port is  used to provide  access to multiple tools.}} &
\multicolumn{1}{c|}{\parbox{2.4cm}{Surgery is carried out  within an anatomical lumen}} &
\multicolumn{1}{c|}{\parbox{2.4cm}{Surgical site accessed  by piercing walls on lumen.}} \\[0pt] \hline
\end{tabular}} \label{tab:surgery_taxonomy}
\end{table}
\par While single port surgery is slowly being adopted in specialty surgical clinics for limited abdominal procedures - mostly carried out trans-umbilically, the clinical acceptance of endoluminal procedures is slow. Both single port and endoluminal procedures are not yet a clinical standard due to lack of clinical evidence supporting the projected patient benefits and due to the technical difficulties in completing these procedures. In part, the lack of clinical evidence is due to lack of commercially available and widely adopted surgical platforms that can help surgeons address these technical challenges while allowing surgeons to focus on optimizing and quantifying surgical outcomes using these new surgical paradigms.   
\section{Challenges to Surgeons During\\ Minimally Invasive and Intraluminal Surgery}
\par
\begin{figure}[htbp]
\center
\includegraphics[width=0.7\textwidth]{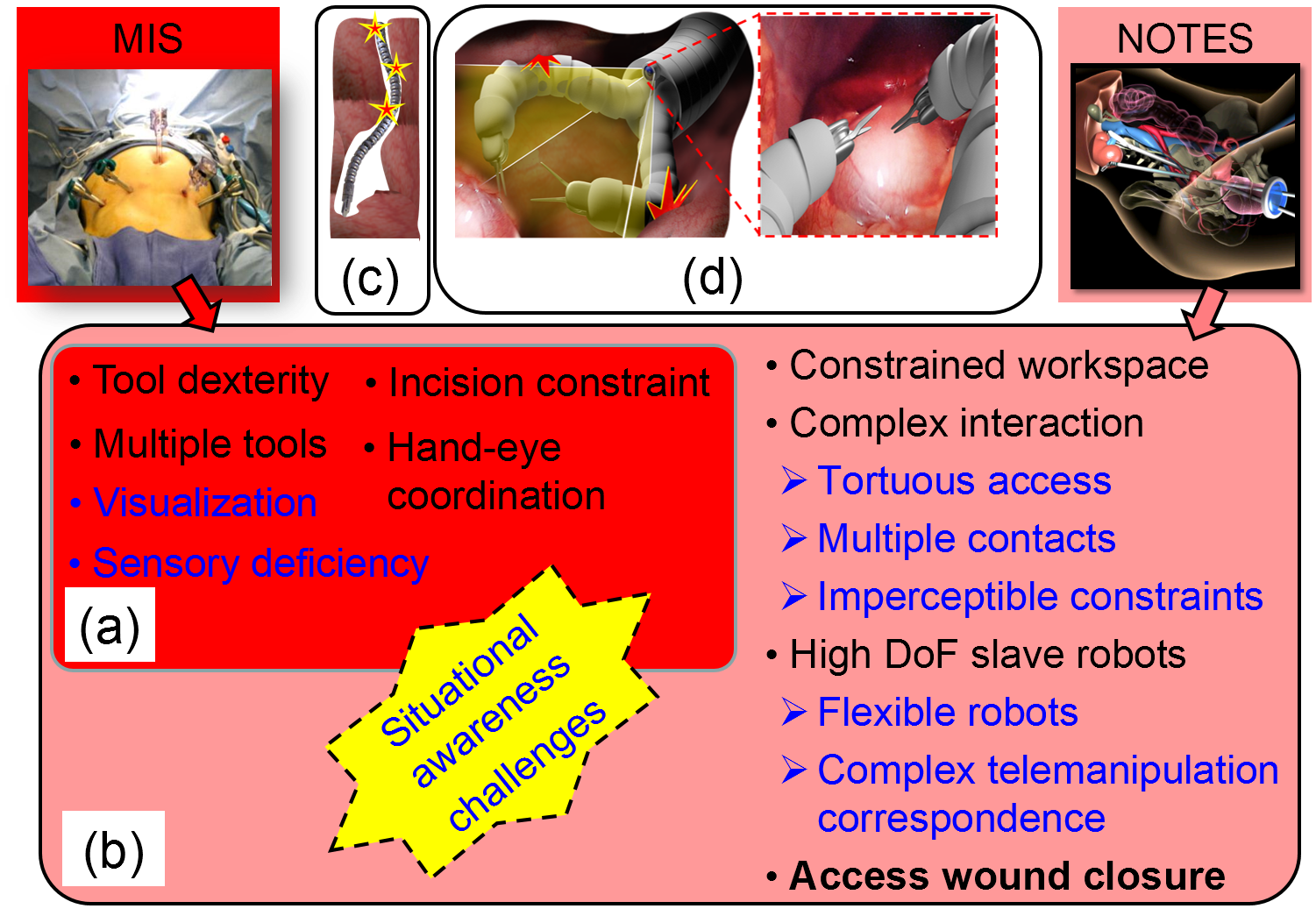}
\caption{The challenges of MIS as a subset of the challenges of NOTES. (a) challenges of MIS, (b) added challenges of intraluminal surgery, (c) multiple contacts along the robot body during endoluminal surgery, (d) an illustrative scenario depicting limited visual perception during intraluminal surgery. The contacts between the robot arm and the anatomy are outside the visualization field of the endoscopic camera. NOTES inset is reproduced with permission from \cite{kaouk2010transvaginal}}
\label{fig:mis_notes_challenges}
\end{figure}

 During minimally invasive surgery, surgeons use robotic assistance to overcome some of the challenges shown in Figure~\ref{fig:mis_notes_challenges}-(a). The first challenge is the need to manipulate several instruments including graspers, suction, endoscopes, clips, and retractors. The second challenge is the abdominal incision point constraint, which requires coordinated motion and reduces the degrees of freedom of rigid instruments to four. This means that surgeons have to overcome the learning curve of manipulating tools inside the body while observing a screen if a stereo-vision display. In addition, visualization through endoscopes provides a narrow view with some depth perception if a large stereo-endoscope is used. This narrow field of view will often require surgeons to readjust their endoscope to be able to visualize the entire surgical site. As a result, surgeons also have difficulty correlating a given endoscope view into an overall chart representing the anatomy of interest. This limitations creates situational awareness challenges and surgeons have to rely of highly specialized training and image-guided navigation aids to overcome these challenges. Finally, the use of surgical tools to interact with tissue instead of the surgeon's hands results in sensory deficiency in terms of feeling forces, texture, temperature and stiffness.
\par Figures~\ref{fig:mis_notes_challenges}(b)-(d) illustrate the encumbered difficulty of single port or intraluminal surgery compared to multi-port MIS. In addition to the challenges of MIS, intraluminal surgery add the complexity of operating in constrained workspace and traversing anatomical passageways. As a result, robots for endoluminal and single port access surgery are required to have highly dextrous architectures with many actuated joints. Also, in procedures such as trans-gastric abdominal surgery there is the significant challenge of obtaining wound closure within the gastric wall after completing the procedure.
\par The unique architecture of intraluminal and single port-access robots present surgeons with significant challenges in terms of perception and situational awareness.  These robots often have to traverse indirect routes of access along tortuous anatomical passageways. As a result, they will contact the anatomy along their entire body, Figure \ref{fig:mis_notes_challenges}-(c). Such contacts will often be outside of the visual perception limits of an endoscopic camera (e.g. see scenario depicted in  Figure~\ref{fig:mis_notes_challenges}-(d)) and, hence, surgeons are oblivious to what occurs outside of their endoscopic field of view. Furthermore, surgeons have to learn how to use the unique architectures of robots for single port access and intraluminal surgery. Often, these robots will have more structural flexibility as a byproduct of having very long bodies as required by the deep location of surgical site relative to the point-of-entry into the anatomy. Also, surgeons have to learn a complex mapping of the joint and workspace limits of such highly dexterous robots so as to be able to complete surgical tasks without causing the robot to assume awkward configurations or hitting the robot's joint limits. In traditional multi-port surgery this problem is less frequent because there is often direct correspondence between the motion of the robot's end effector and the wrist of the surgeon manipulating via a surgical console. Such intuitive mapping is often not possible when using dedicated robots for NOTES and intraluminal surgery.

\section{Snake-Like Robots for MIS, SPA and NOTES}
\par The added constraints of SPA and intraluminal surgery place higher demands for robots that can provide distal dexterity in confined spaces while being able to triangulate at least two arms to facilitate dual-arm tissue approximation and suturing as in open-surgery. Concurrent developments enabling miniature camera technology have also been critical for advancing new miniature insertable visualization aids such as \cite{hu_allen2009insertable_camera, Gu2015}. With the visualization challenges solved, the last decade has seen a flurry of research activity and new designs of snake-like robots and systems for SPA and intraluminal surgery. This section reviews the mechanical architectures of surgical snake-like robots, identifies design and functional requirements and provides examples of state-of-the art surgical systems for SPA and intraluminal surgery.
\subsection{Mechanical Architectures of Snake-like Robot }
\par Table~\ref{table:snake_robots_architectures} on page~\pageref{table:snake_robots_architectures} shows an overview of research platforms and robotic systems using snake-like robots for SPA and intraluminal surgery. A current review of systems for SPA and NOTES was recently presented in \cite{Haber2010, Marcus2014, zhao_Xu2015NOTES_review, arkenbout2015_notes_review}. The table presents surgical systems based on the backbone type for their snake-like arms. There are three backbone types: continuous, discrete and hybrid. Robots with continuous backbones (often referred to as continuum robots) use a continuous elastic backbone that is bent by wires, push-pull actuation or by antagonistic pairs of pre-shaped superelastic tubes. Robots with discrete backbones use articulated linkages, pivots and wire-compressed cams to form their structure. Hybrid backbone robots use a mixture of flexible elements (e.g. springs) and linkages to achieve manipulation.
\par
\begin{figure}[htbp]
\center
\includegraphics[width=0.95\textwidth]{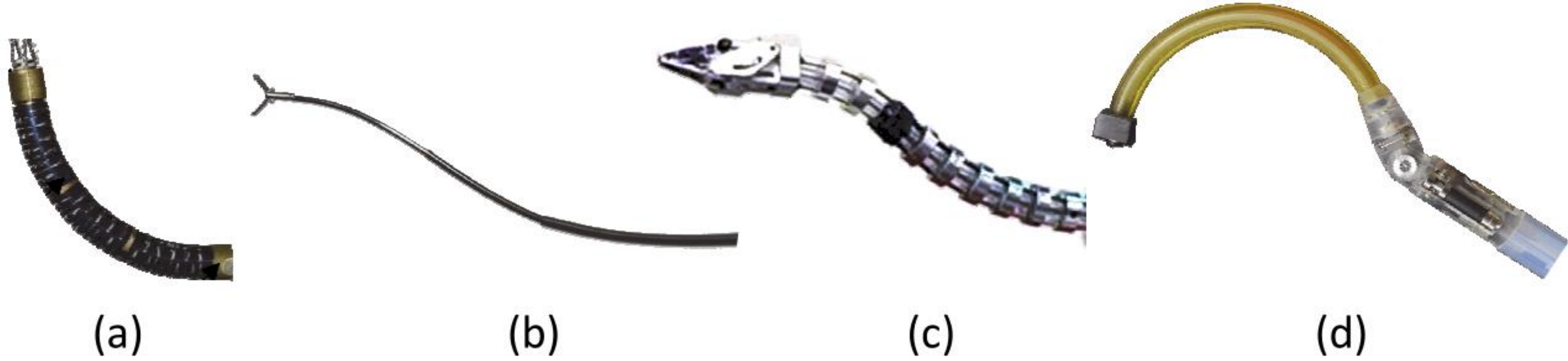}
\caption{Continuum robots for surgery: (a) a flexure-based single backbone robot with wire actuation \cite{peirs2003flexure_snake}, (b) a single backbone robot concentric tube robot (courtesy of R. Webster \cite{webster2006toward}), (c) a two-segment multi-backbone continuum  robot \cite{Simaan2009}, (d) interleaved continuum robot\cite{conrad2016interleaved}}
\label{fig:concentric_tube}
\end{figure}

 Several reviews of continuum robots and their applications have been presented in \cite{webster_jones2010_continuum_review, Burgner-Kahrs2015, gilbert2016concentric}. There are generally three types of continuum robots used for surgical applications: wire-actuated single backbone continuum robots (e.g. \cite{peirs2003flexure_snake, kutzer2011design}), single backbone concentric tube robots or active cannulas (e.g. \cite{webster2006toward, sears2006steerable, furusho2005development}) and multi-backbone continuum robots (MBCRs) (e.g. \cite{Simaan2004, Simaan2009}).  Figure~\ref{fig:concentric_tube}-(a) shows an early example of a single segment continuum robot having a backbone made of superelastic $\diameter5$ mm Nickel-Titanium (NiTi) tube equipped with flexures. This device can bend in two degrees of freedom using four wires\cite{peirs2003flexure_snake}. Figure~\ref{fig:concentric_tube}-(b) shows an example of a single backbone concentric tube robot. Such robots are typically formed using antagonistic pairs of pre-bent superelastic NiTi tubes. By sliding and rotating these tubes relative to each other 3D equilibrium shapes may be obtained. Figure~\ref{fig:concentric_tube}-(c) shows an example of a multi-backbone continuum robot. This type of robots uses several backbones in push-pull actuation to achieve bending of snake-like segments. A hybrid articulated-continuum robot is shown in  Figure~\ref{fig:concentric_tube}-(d). This design was recently introduced to further increase dexterity and to expand the design space for surgical robots. The design uses wire-actuated wrists with wire-actuated segments\cite{conrad2016interleaved}. Though it provides continuum robots with hinged rotation among its segments, the wire actuated wrists enabling this rotation have limited torque capabilities.
\begin{table}[H]
\centering
\tbl{Mechanical Architectures of Snake-Like Surgical Robots\vspace{0.5\baselineskip}}{
\label{table:snake_robots_architectures}
\fontsize{12}{6mm}\selectfont
\resizebox{1.0\columnwidth}{!}
{
\begin{tabular}{| >{\centering\arraybackslash}m{2cm}| >{\centering\arraybackslash}m{2.3cm}| >
{\centering\arraybackslash}m{3.5cm} | >{\centering\arraybackslash}m{6.5cm} | >{\centering\arraybackslash}m{2cm}|}

\hline
\textbf{Backbone Type} & \textbf{Actuation Type} & \textbf{Actuation Transmission} & \textbf{Sample Works} & \textbf{Wrist }\\ \cline {1-5}

\multirow{13}{2cm}{\centering Continuous} & \multirow{9}{2.3cm}{\centering Electro- mechanical}  & Pre-curved tubes & Transurethral prostate surgery \cite{Hendrick2014,Mitchell2016} & - \\[1mm]  \cline {3-5}
&  & \multirow{3}{2.5cm}{Wire-driven} & Monolithic multi-arm structure \cite{Roppenecker2014} & - \\[1mm]  \cline {4-5}
&  & & STRAS \cite{DeDonno2013} & -\\[1mm]  \cline {4-5}
&  & & Transenterix Surgibot \cite{walker2016future} & Roll \\[1mm]  \cline {3-5}
&  & & Transoral multi-backbone system \cite{Simaan2009} & Backbone rotation \\[1mm]  \cline {4-5}
&  & \multirow{1}{2cm}{Push-pull} & IREP \cite{Ding2013,Bajo2012} & Roll \\[1mm]  \cline {4-5}
&  &  & SURS \cite{Xu2014} & - \\[1mm]  \cline {4-5}
&  &  & Multi-backbone snake for NOTES \cite{Zhao2013} & - \\[1mm]  \cline {2-5}
& \multirow{2}{2.3cm}{\centering Manual} &\multirow{2}{2.5cm}{\centering Wire-driven}& Transenterix SPIDER \cite{Haber2012}
& Roll \\[1mm]  \cline {4-5}
&  &  & Anubis \cite{Dallemagne2010} & -  \\[1mm]  \cline {2-5}
& Pneumatic & Bellows & STIFF-FLOP \cite{Cianchetti2013,Shiva2016} & - \\[1mm]  \cline {1-5}

\multirow{19}{2cm}{\centering Discrete} & \multirow{17}{2.3cm}{\centering Electro- mechanical} &  \multirow{4}{2cm}{\centering Linkage} & Double Screw Drive mechanism \cite{Liu2013} & -  \\[1mm] \cline {4-5}
& &  & Stackable 4-bar manipulator \cite{Lee2012} & Yaw \\[1mm] \cline {4-5}
&  &  & Transluminal magnetic modules  \cite{Tognarelli2015,Tortora2013} & - \\[1mm] \cline {4-5}
&  &  & Dual parallel robots \cite{Matich2015,Matich2015a} & -  \\[1mm]\cline {3-5}
&  & \multirow{11}{2.5cm}{\centering Wire-driven} & Rigid arms with distal dexterity \cite{Seung2011} & RPY\\  \cline {4-5}
&  &  & SAIT Surgical Robot \cite{Lee2014,Kim2014,Roh2015_choi_SPA_SAIT} & PYR  \\[1mm]  \cline {4-5}
&  &  & Rigid arms with distal dexterity \cite{kwon2013_single_port_RCM_linkage} & PY \\[1mm] \cline {4-5}
&  &  & HARP / Flex Robotic System \cite{Ota2008,Rivera-Serrano2012,Lang2017} & -  \\[1mm] \cline {4-5}
&  &  & ViaCATH \cite{Abbott2007} & PY \\[1mm] \cline {4-5}
&  &  & micro-IGES for TEMS \cite{Shang2017} & PY \\[1mm]  \cline {4-5}
&  &  & Interlocking joints \cite{Liu2016} & -  \\[1mm]  \cline {4-5}
&  &  & HVSPS single-port system \cite{Can2012} & Roll  \\[1mm]  \cline {4-5}
&  &  & Pelvic access robot \cite{Clark2015} & -  \\[1mm] \cline {4-5}
&  &  & da Vinci Sp \cite{Kaouk2014} & -  \\[1mm] \cline {3-5}
&  &  \multirow{2}{2.5cm}{\centering Push-pull} & SPRINT \cite{Piccigallo2010} & RPR  \\[1mm]\cline {4-5}
&  &  & MASTER \cite{Phee2009} & RY \\[1mm] \cline {2-5}
& SMA & Wire-driven & Laparoscopic arm with wrist \cite{Yuan2014,Shi2016} & Spherical  \\[1mm] \cline{1-5}

\multirow{7}{2cm}{\centering Hybrid} & \multirow{7}{2.3cm}{\centering Electro- mechanical}  & Wire-driven & Spring and ball-joint backbones for maxillary sinus surgery \cite{Yoon2013} & -  \\[1mm] \cline{3-5}
&  & Wire-drive and concentric tube & Highly articulable probe with concentric tubes\cite{Mahvash2011} & - \\[1mm]  \cline {3-5}
&  & Tendons / discrete actuators & Interleaved continuum-rigid manipulation\cite{conrad2016interleaved}  & Backbone rotation \\[1mm] \cline{3-5}
&  & Linkage / plate spring  & PLAS \cite{cheon2014_Hong_DGIST_SPA} & Roll  \\[1mm] \cline {4-5} \hline

\end{tabular}
}}

\begin{flushleft}
$\dagger$ Wrist DOF are denoted as ``R'' for roll, ``P'' for pitch, and ``Y'' for yaw, in order starting at the base of the wrist. ``Backbone Rotation'' refers to rolling the snake's body about its backbone.
\end{flushleft}

\end{table} 
\par Segments of multi-backbone and wire-actuated single backbone robots can actively control their radius of curvature without a change in their length. Concentric tube robots, on the other hand, necessitate a change of the length of their tube pairs to achieve a specific bending radius. Concentric tube robots are generally suitable for delicate needle-type operations or surgical tasks involving small forces (generally smaller than 1 N). In comparison, MBCR's and wire-actuated single-backbone robots can carry larger loads at the cost of larger diameters.
\par Other types of snake-like robots for surgery are directly inspired by designs for surgical wrists. A review of technologies used for design of surgical wrists was provided by Jel\'inek et al. \cite{Jelinek2015classification_wrists_Breedveld}. The designs of snake-like robot for  SPA  and Intraluminal surgery  include wire actuated articulated designs \cite{Roh2015_choi_SPA_SAIT, cooper2004_davinci_snake}, linkage-based designs\cite{yamashita2003multi_slider,  Lee_SAIT_SPA_Choi2014, Lee2010_Byung_Ju_Yi_stackable_fourbar, cheon2014_Hong_DGIST_SPA}, differential screws  \cite{ishii2010_screw_kobayashi}, wire-actuated rolling cams \cite{Lee_SAIT_SPA_Choi2014, cooper2004_davinci_snake},  wire-actuated with spherical spacers and disks \cite{harada2005_2.5mm_spheres_fetal_surgery_fujie} or sliding hemispherical links \cite{sturges1993flexible, sturges1996voice, degani2006_harp_choset}, wire-actuated constrained linkages \cite{kwon2013_single_port_RCM_linkage, Jel2013_Dragon_flex_Breedveld}.
\subsection{Specific Design and Functional Requirements for Single Port and Intraluminal Surgery}
\par Single port access surgery (SPA) and intraluminal surgery place special constraints on the designers. These constraints must be addressed when designing new robotic systems for successful deployment. For example, intraluminal and SPA surgery often require multiple robotic arms to operate though a narrow access channel or anatomical passageways. This will often require careful consideration of how the actuation units for each one of the robotic arms in designed and mounted in order to avoid collision between them. Also, the workspace constraints often mean that the robotic arms have to emanate from a narrow access over-tube; thus, placing strict constraints on kinematic dexterity and workspace. To overcome this challenge, Ding et al. \cite{ding2010design} have considered the effect of designing a dedicated mechanism for changing the distance between the bases of the two snake-like arms of an SPA robot. This work showed a distinct advantage of having an adjustable distance of the "elbows" of a dual-arm robot for SPA in terms of dexterity. While many research systems for SPA do not follow this principle (e.g., \cite{Simaan2009, DeDonno2013, Mylonas2014}), recent designs allow some control of the baseline distance between the "elbows" of the robotic arms (e.g., \cite{Ding2013, xu2015_unfoldable_spa, Lee_SAIT_SPA_Choi2014} and da Vinci's recent single port access system).
\par
\begin{figure}[htbp]
    \centering
    \centerline{\includegraphics[width=0.9\columnwidth]{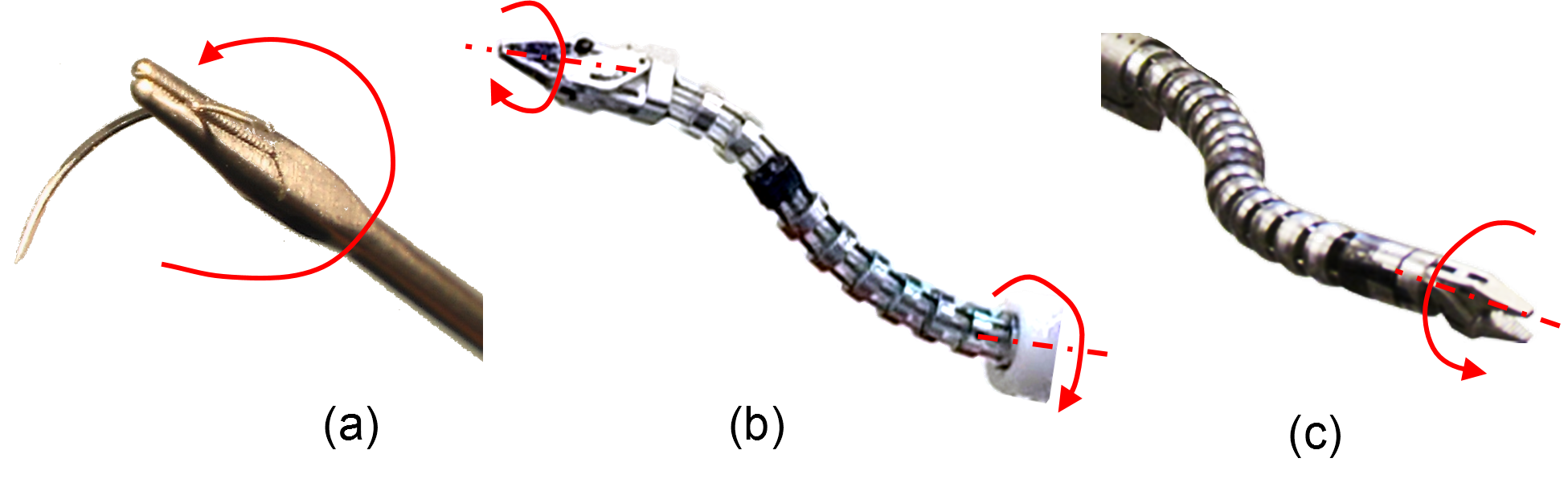}}
    \caption{Suturing in confined spaces: (a) axial rotation about a trans-oral laryngologist needle holder, (b) gripper axial rotation through transmission of rotation about the backbone, (c) gripper axial rotation through a dedicated wrist\cite{ding2010design} }
    \label{Figure:suturing_alternatives}
\end{figure}

 Another key capability for successful minimally invasive surgery is tissue reconstruction using suturing. Figure~\ref{Figure:suturing_alternatives}-(a) shows one of the key modes of operation by which surgeons pass suturing needles in confined spaces. Instead of using straight needles, surgeons use circular needles and rotation of a needle holder about its longitudinal axis to pass the needle. The success of SPA systems and intraluminal systems in achieving effective tissue reconstruction following excision depends on the abilities of their dexterous arms to replicate this mode of operation. Figure~\ref{Figure:suturing_alternatives}-(b) shows how multi-backbone continuum robots can be used for rotation transmission to facilitate passing circular needles for suturing in confined spaces \cite{Simaan2009}. This mode of operation however requires significant modeling and characterization of actuation losses in order to achieve stable rotation of the gripper about its longitudinal axis.  To overcome this challenge, it is possible to use the solution shown in Figure~\ref{Figure:suturing_alternatives}-(c) where a dedicated roll wrist can facilitate the passing of circular needles. As a testimony to the importance of gripper roll about its longitudinal axis one can observe that  Table~\ref{table:snake_robots_architectures} shows a list of several systems with wrists having their last joint be a roll joint.
\par The above two key issues are not the only challenges that designers must address when considering new design concepts for systems geared towards SPA or intraluminal surgery. The list includes special considerations for safety and and ease of deployability of such robots into the patient, serialization and maintainable of insufflation, visualization of the surgical field, introduction and removal of clips and surgical consumable to and from the surgical site, extraction and bagging of excised tumors despite the narrow access. Many of these issues remain partly or not suitably addressed  in existing systems for SPA and intraluminal surgery.

\subsection{Systems for Single Port Access Surgery}
\par Single Port Access (SPA) is  a minimally invasive surgical technique that enables surgeons to operate on internal organs via a single orifice in the body. This technique is mostly suitable for abdominal surgeries with the navel as the entry port, resulting in minimal scarring. During SPA procedures, the surgeon inserts and manipulates the cutting, visualization, and insufflation instruments through one port. However, operating with straight tools through a single incision results in instrument collision and significantly reduces the surgeon's workspace due to loss of triangulation. An additional challenge is that the crossing of instruments causes a reversal of handedness and increases difficulty in hand-eye coordination.
\par The needs of SPA surgeries are not fully met by existing minimally invasive multi-port surgical platforms. These platforms are generally unsuitable for single port access due to their inability to align their tools through a single trocar/access port while avoiding collision between the actuation units of each arm. The rest of this section reviews research developments that attempt to address the needs of single port surgery. These systems have indirectly informed the design of existing and upcoming commercial systems. Prior to delving into details about some of these systems, we point the readers to Table \ref{table:systems_for_SPA_surgery} which presents a summary of existing robotic systems for single-port surgery.
\begin{table}[H]
\centering
\tbl{Summary of existing systems for single-port surgery\vspace{0.5\baselineskip}}{
\label{table:systems_for_SPA_surgery}
\fontsize{9}{5mm}\selectfont
{
\begin{tabular}{| >{\centering\arraybackslash}m{3.8cm}| >{\centering\arraybackslash}m{2.8cm}| >
{\centering\arraybackslash}m{1.8cm} | >{\centering\arraybackslash}m{2.0cm}|}\hline

\textbf{System or Developer} & \textbf{Port Size (mm)} & \textbf{Arm DoFs} & \textbf{Payload (N)}\\ \cline {1-4}
Da Vinci SP system\cite{Kaouk2014}&\diameter 25 & 7 & -\\ \cline {1-4}
IREP \cite{Ding2013} & \diameter 15 & 7 & - \\ \cline {1-4}
SURS\cite{xu2015_unfoldable_spa}  & \diameter 12 & 6 & 2 \\ \cline {1-4}
Lee et al.\cite{Lee2010_Byung_Ju_Yi_stackable_fourbar} & \diameter 25 & 5 & - \\ \cline {1-4}
SPRINT \cite{Niccolini2012}  & \diameter 30 & 6 & 5 \\ \cline {1-4}
SISR \cite{Wortman2013}& \diameter 30 & 4 & - \\ \cline {1-4}
SAIT\cite{Kim2014,Lee2013} & \diameter 30 & 7 & - \\ \cline {1-4}
Kobayashi et al.\cite{Liu2013,kobayashi2015}& \diameter 25 & 6 & - \\ \cline {1-4}
Kwon et al.\cite{kwon2013_single_port_RCM_linkage} &  \textgreater \diameter 16 & 6 & \textgreater 7.5 \\ \cline {1-4}
PLAS Surgical Robot  \cite{cheon2014_Hong_DGIST_SPA}&  \diameter 25 & 6 & \textgreater 14 \\ \cline {1-4}
\end{tabular}
}}
\begin{flushleft}
$\dagger$ This table was adapted from Xu et al.\cite{xu2015_unfoldable_spa} with the author's permission.
\end{flushleft}

\end{table}

\par
\begin{figure}[htbp]
    \centering
    \centerline{\includegraphics[width=0.95\columnwidth]{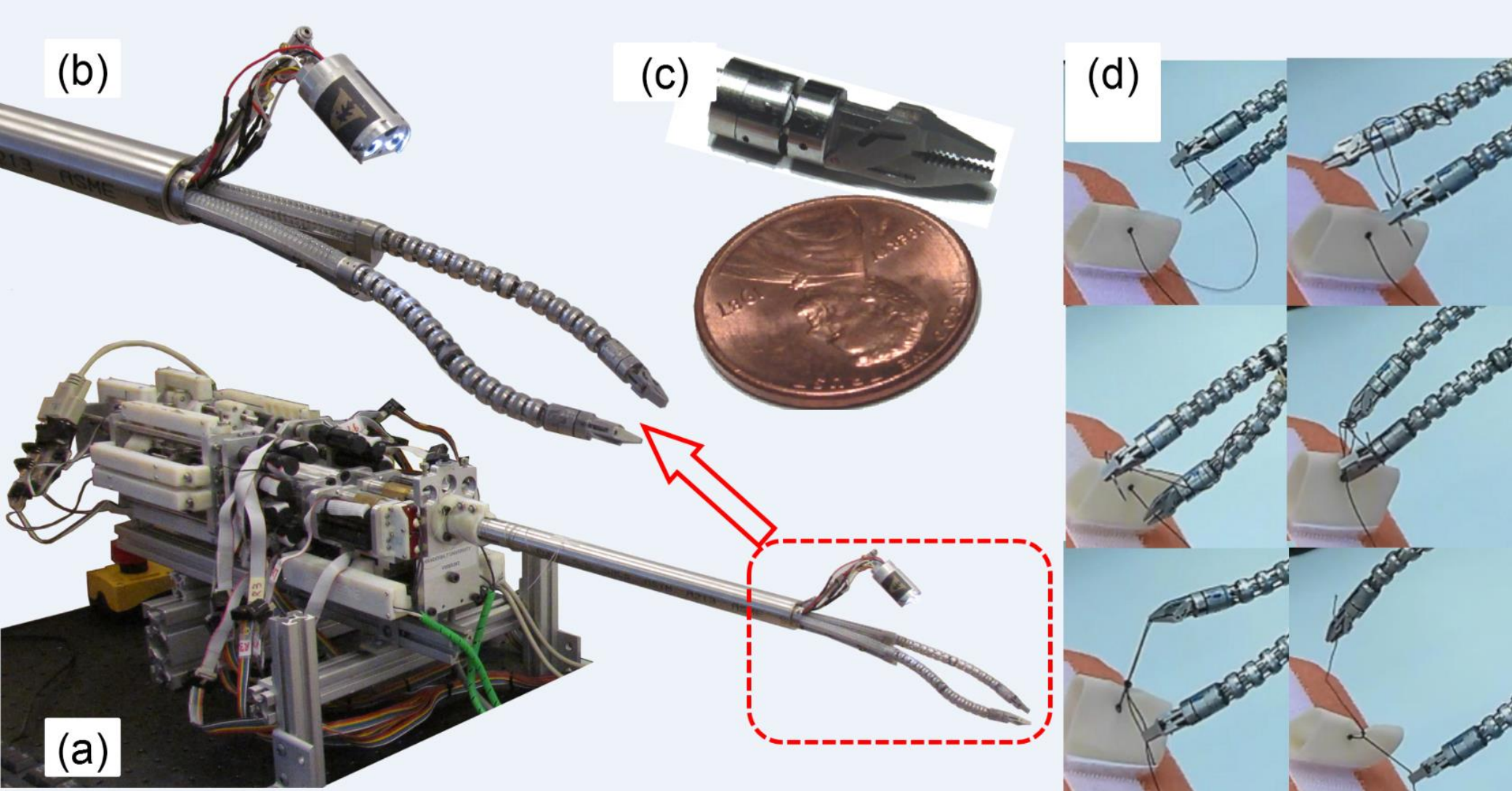}}
    \caption{(a) The IREP single port access system, (b) the stereo-vision head with two seven degrees of freedom arms, (c) the distal wrist and gripper, (d) example of tying a double-throw knot. \cite{Ding2013}}
    \label{Figure:IREP}
\end{figure}

 The Insertable Robot Effectors Platform (IREP), shown in Figure \ref{Figure:IREP}, was developed to operate through the abdomen via  a $\diameter 15$ mm port\cite{Xu_IREP_IROS2009}. The IREP is equipped with two $\diameter 6.4$ mm 5-DoF snake-like continuum robots, two 2-DoF parallelogram mechanisms, and one 3-DoF stereo vision module. The two continuum arms of the IREP are equipped with a wrist that provides roll along its longitudinal axis, which is particularly advantageous for suturing in confined spaces. The IREP enters the abdomen in its folded configuration with its stereo-head display pointing forward. It then self-deploys into its working configuration once at the target site of surgery. The parallelogram linkages enable intra-abdominal triangulation, and enables the surgeon to reach a workspace volume of about 125 cm$^{3}$, which corresponds to the size of the surgical site on target organs\cite{Ding2013}. A preliminary evaluation of the IREP showed that it is capable of providing 0.24 mm root mean squared error of path following when tele-manipulated by users using high-definition vision feedback to the users\cite{simaan2013lessons_IREP_j_robotic_surgery}. Among its noted limitations was the range of roll about the longitudinal axis of its gripper. Initially designed with $\pm 60^\circ$ roll about the gripper axis, it was found through suturing experiments that this range should ideally have been larger due to coupling between the continuum arm motion and the orientation of the wrist. The IREP has been licensed to Titan Medical and is being commercialized as the Titan SPORT\textsuperscript{TM} Surgical System.
\par
\begin{figure}[htbp]
    \centering
    \centerline{\includegraphics[width=0.6\columnwidth]{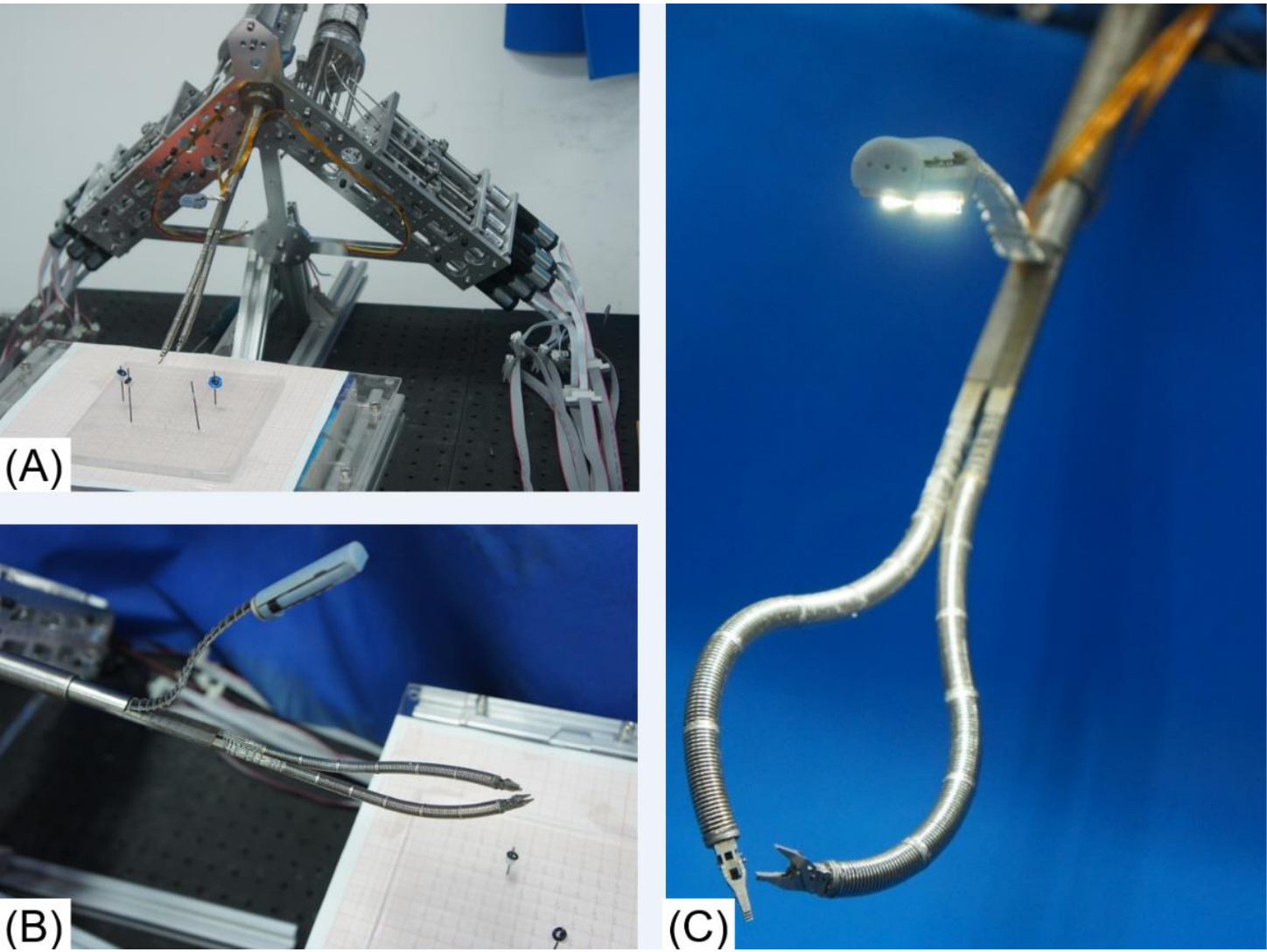}}
    \caption{A triple-arm robot using two continuum arms for manipulation and one for manipulating a stereo vision head.   \cite{xu2015_unfoldable_spa}}
    \label{Figure:kai_single_port}
\end{figure}

 The SJTU Unfoldable Robotic System (SURS), displayed in Figure \ref{Figure:kai_single_port}, consists of two 6-DoF multi-segment continuum end effectors and a 3D vision unit. The system enters the abdominal wall in its folded configuration through a 12 mm port, and then unfolds into its working configuration. The vision unit, coupled with illumination, was designed to have a cylindrical shape in order to facilitate insertion. The end effectors of this manipulator have payload capabilities of 2 N. In addition, SURS robot is highly modular; its end effectors can easily be replaced by grippers, needles, or ablations tips during surgery. The base of the robot is attached to a 6R industrial robot, which serves as a remote center of motion (RCM) mechanism for coarse positioning of the SURS about abdominal point of entry\cite{xu2015_unfoldable_spa}. One of the limitation of the SURS system is that it does not possess a distal rotary wrist.
%
 %
\par
\begin{figure}[htbp]
    \centering
    \centerline{\includegraphics[width=0.9\columnwidth]{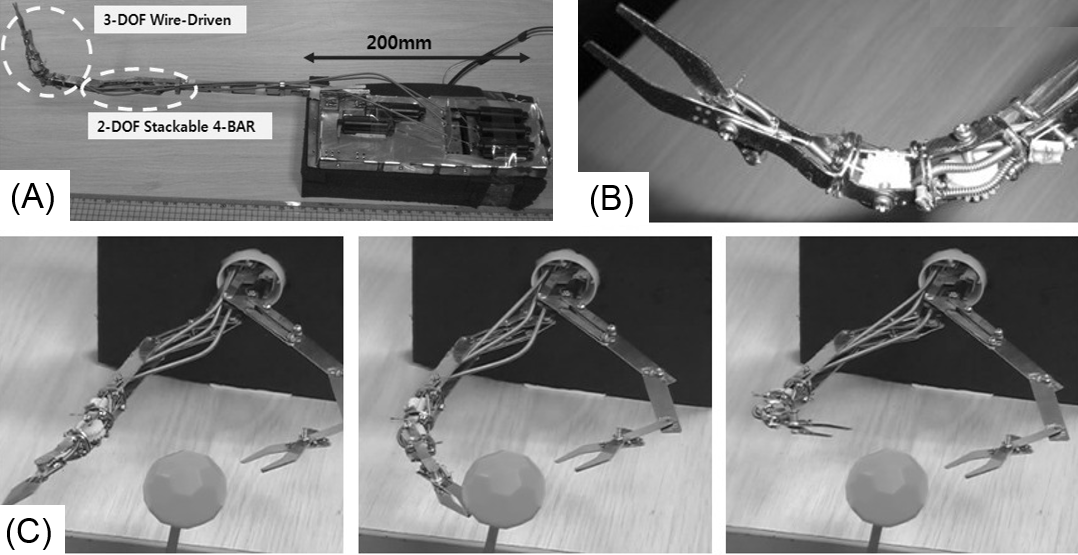}}
    \caption{(a) A stackable four-bar manipulator for SPA, (b) close-up view of 3-DOF wire-driven end effector, (c) demonstration of insertion through a 25 mm port. \cite{Lee2010_Byung_Ju_Yi_stackable_fourbar}}
    \label{Figure:bj_fivebar}
\end{figure}

 Another design approach for deployable SPA robots was presented by Lee et al.\cite{Lee2010_Byung_Ju_Yi_stackable_fourbar}. The design uses stackable 4-bar manipulators robot for SPA surgery, through a $\diameter 25$ mm port. This design, shown in Figure \ref{Figure:bj_fivebar}, has the advantage that each of its 5-DOF arms can be detached from the actuators, rendering the system modular and lightweight. In addition, each joint is driven by a separate 4-bar mechanism, which increases the robustness of the system\cite{Lee2010_Byung_Ju_Yi_stackable_fourbar}.

\par
\begin{figure}[htbp]
    \centering
    \centerline{\includegraphics[width=0.7\columnwidth]{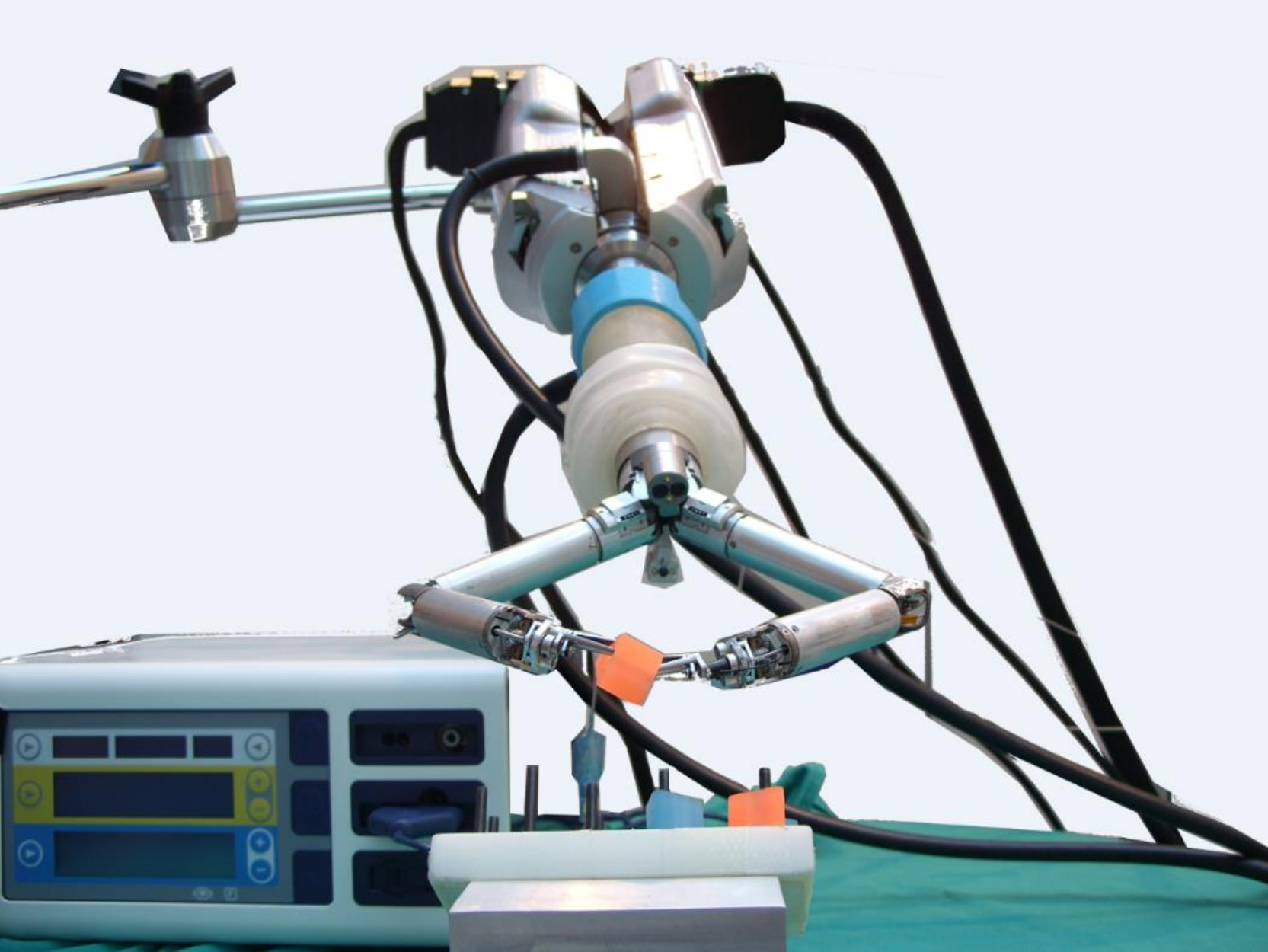}}
    \caption{The SPRINT robot demonstrating dual-arm action with a stereo vision head. \cite{Niccolini2012}}
    \label{Figure:sprint_arianna}
\end{figure}

 The European ARAKNES project developed a Single-Port lapaRoscopy bimaNual roboT (SPRINT), shown in Figure \ref{Figure:sprint_arianna}. The SPRINT manipulator consists of two 6-DoF anthropomorphic robotic end effectors with payload capabilities of 5 N. These miniaturized arms replicate the surgeon's motions in real-time via master-slave teleleoperation via a dedicated console. This design embeds electromechanical motors inside its $\diameter 23$ mm surgical arms and requires a $\diameter 30$ mm access port. The joint distribution on each arms consist of a Roll-Pitch-Pitch serial chain, followed by a Roll-Pitch-Roll wrist motion. In addition to the two primary arms, up to two additional arms of smaller size can be inserted, to hold additional instruments such as a telescopic-camera holder. This modular design enables surgeons to switch instruments on the fly. Furthermore, this robotic platform was designed with an open 12 mm central lumen, where assistive tools such as hemostatic sponge or suturing needles can be inserted as needed\cite{Niccolini2012}.

%
\par The single-incision \textit{in vivo} surgical robot (SISR), developed by Wortman et al.\cite{Wortman2013}, is the first entirely \textit{in vivo} robot for single port colectomy. The SISR is a bimanual system, with each 4-DOF arm comprised of a torso, upperarm, and forearm. The end effectors of this manipulator are equipped with a rotational degree-of-freedom, as well as an open/close actuation unit. During surgery, the arms are straightened, separated, and inserted individually via a single 30 mm abdominal incision. They are then mated and reassembled within the abdomen using control rods attached to the torso segments. To facilitate this insertion and prevent collision with internal organs, Army-Navy retractors are used to lift the abdominal walls. All of the actuation motors of the SISR are contained within the abdominal cavity. This design parameter removes some of the constraint of the entry port, allowing the robot to be easily repositioned during surgery on large organs.

\par
\begin{figure}[htbp]
    \centering
    \centerline{\includegraphics[width=1.0\columnwidth]{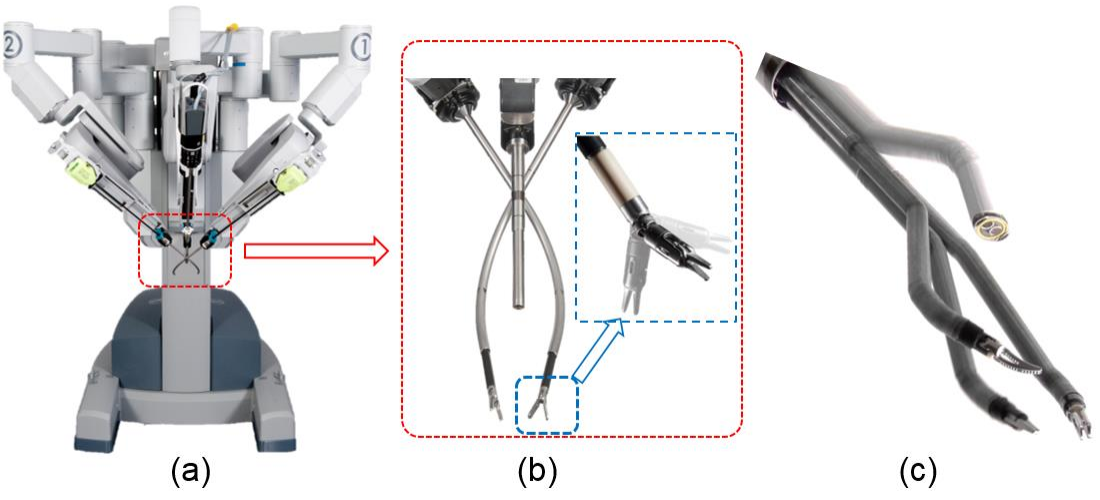}}
    \caption{(a) The Da Vinci Single-Site Platform (DVSSP), (b) Close-up of arms showing two degrees of freedom wrists, (c) The Da Vinci Single Port Surgical System}
    \label{Figure:davinci_single_port_and_site}
\end{figure}

 Intuitive Surgical's da Vinci Single-Site\textsuperscript{\textregistered} Platform (DVSSP), shown in Figure \ref{Figure:davinci_single_port_and_site}-(a), was the first robotic system approved for SPA surgery by the Food and Drugs Administration (FDA). This platform consists of a set of single-site surgical tools, including two robotically-controlled curved canulae, two straight cannulae, and an insufflation valve. The DVSSP tools penetrate the abdominal wall through a 25 mm port. Although it was originally approved for single site cholecystectomy  \cite{kroh2011first}, the DVSSP has also been adapted for uses in other types of abdominal surgeries, and well as urology, and gynaecology\cite{morelli2015vinci}. According to a study conducted by Morelli et al, the principal reported advantage of the DVSSP has been the restoration of intra-abdominal triangulation. The DVSSP achieves this triangulation by crossing its curved instruments midway through the entry port. In addition, the surgeon's hand-eye coordination is improved through reassignment of the tools' kinematic mapping in the da Vinci system's software. Nevertheless, the DVSSP has been criticized for lacking the distal wrist action 
that is essential for robot-assisted suturing \cite{morelli2015vinci}.
\par A two degrees of freedom wrist was later incorporated, as shown in Figure \ref{Figure:davinci_single_port_and_site}-(b). This wrist provides pitch and yaw motions but is unable to transmit independent roll along its longitudinal axis. The new edition of the DVSSP can achieve suturing, but only through a complex combination of pitch and yaw motions.
\par In 2014, a second SPA robot was approved by the FDA: Intuitive Surgical's da Vinci Sp Surgical System\cite{Kaouk2014}, shown in Figure \ref{Figure:davinci_single_port_and_site}-(c). This robot is equipped with three $\diameter 6$ mm articulating endoscopic arms and an endoscopic camera. The Sp system is inserted in the patient through as single port as a single cylindrical robotic tool. Once at the site of surgery, the arms and camera are deployed from the robotic port and controlled independently. This single-port system can be equipped with distal wrists that provide pitch and yaw motions. The wrists do not have an independent roll degree-of-freedom, but the dexterity of the robotic arms allows approximate rolling motion about their backbones.
\par A single-port surgical robot being developed by the Samsung Advanced Institute of Technology (SAIT) consists of an articulated guide tube and two wire-driven 6-DoF instruments composed of rigid links and a 3-DOF endoscope that are deployed through the guide tube. The guide tube, which has a diameter of 30 mm, is positioned using a slave robot arm with a remote-center-of-motion mechanism. The guide tube, in addition to having 4-DOF, can adjust its stiffness by changing the pre-tension in the wires due to its variable neutral line design\cite{Kim2014}. Each 6-DOF instrument has a pitch-yaw-roll wrist and pulleys at the joints that reduce actuation wire tension. This robot was able to manipulate a 1 kgf load, but as with any wire-driven robot, friction in the actuation lines reduces efficiency and presents modeling difficulty\cite{Lee2013}.
\begin{figure}[htbp]
    \centering
    \centerline{\includegraphics[width=0.90\columnwidth]{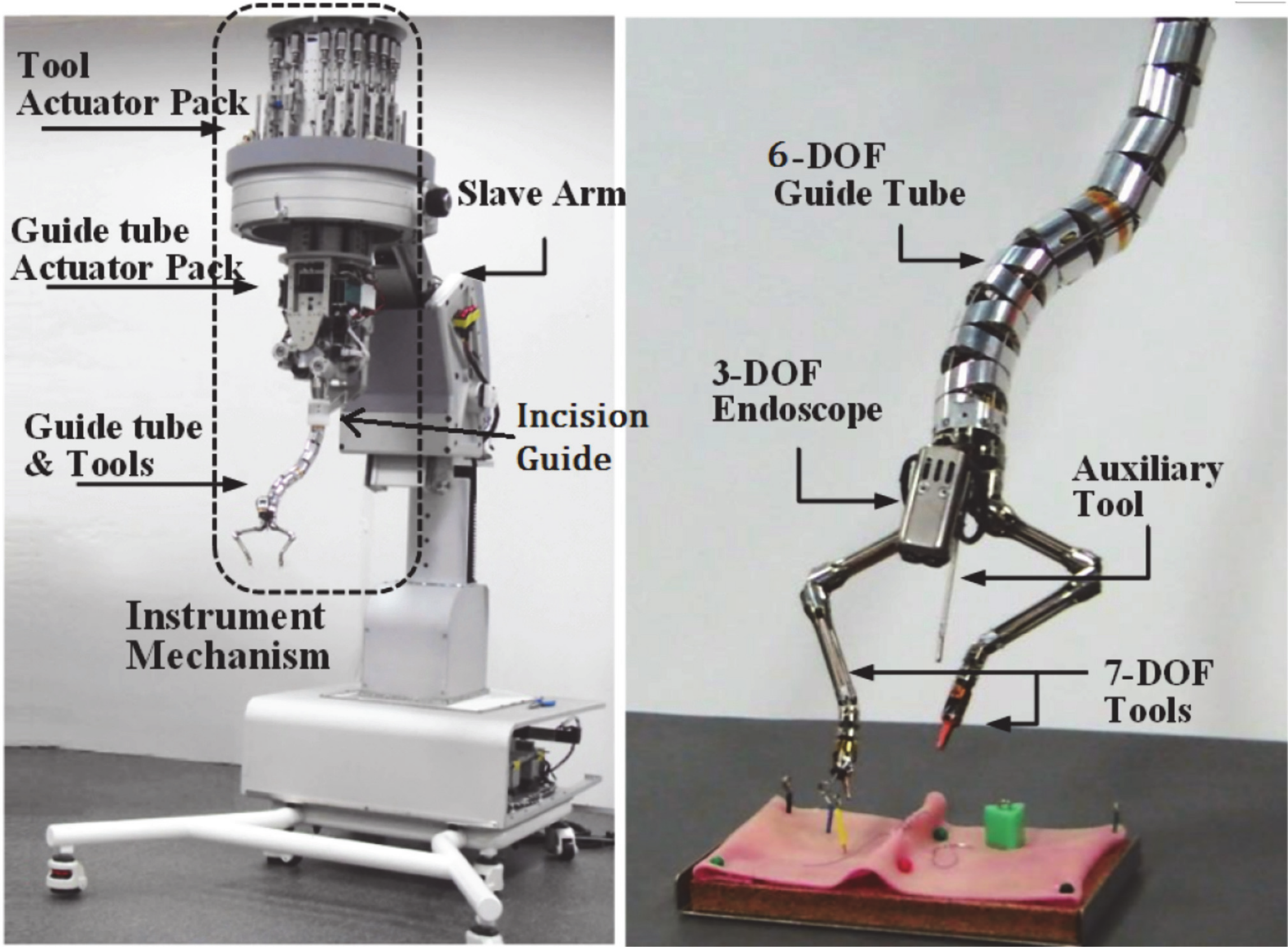}}
    \caption{A prototype SPA system developed by Samsung Advanced
Institute of Technology (SAIT) and Korea Institute of Science and Technology (KIST) \cite{Lee2014}}
    \label{Figure:Lee_choi_SAIT_SPA_robot}
\end{figure} 
\subsection{Systems for Intraluminal Surgery}
\par Intraluminal surgical techniques, including both transluminal and endoluminal techniques, can potentially improve patient outcomes, however, as described previously these methods present multiple technical challenges. There have been recent research efforts to develop robotic systems that will aid surgeons in performing these intraluminal surgeries, which could allow for improved results and increased adoption of intraluminal surgical techniques. Current robotic research includes transnasal, transoral, transurethral, and transanal surgery as target applications.
\subsubsection{Transnasal Systems}
\par Systems for transnasal surgery have been investigated with several exploratory surgical procedures in mind. These procedures ranked in increased distance from the entry point include trans-nasal navigation of the sinuses and biopsy, transnasal skull base surgery, and transnasal micro-surgery of the upper airways.
\par One of the challenges with diseases of the sinuses is the difficulty in monitoring disease progression, obtaining biopsy and offering surgical intervention in the frontal and maxillary sinuses while avoiding visible scarring or obliteration of bone scaffolds of the nose. To address these needs, a dual-arm wire-driven robot for transnasal navigation and biopsy of the sinuses has been presented by Yoon et al. \cite{Yoon2011,Yoon2013}. This robot is comprised of one 5-Dof arm ($\diameter 5$ mm diameter) for biopsy with a 1-DoF gripper, and one 4-DoF steerable endoscope arm ($\diameter 4$ mm diameter) for inspections into the sinuses. The biopsy tool has a central backbone composed of discrete ball-joints for increased stiffness. The inspection arm is a two-segments wire-driven endoscope with a spring backbone and discrete disks along the backbone. The two arms are designed to fit within a diameter of $\diameter 10$ mm, in order to closely match the anatomical constraints imposed by the size of nostrils. This system has been tested on a soft sinus phantom, and the results showed that the biopsy end effector is able to resist pulling forces up to 2.5N.
\par An interesting approach for skull base tumor surgery is through transnasal access. A typical target for these surgeries is the removal of pituitary gland tumors through a transsphenoidal approach. Though first described in 1906 by Scholffer\cite{schloffer1906transnasal_pitutary_first_work}, the manual endoscopic approach for these surgeries are limited by narrow access, cumbersome manual manipulation of surgical tools near very sensitive anatomy and lack of distal dexterity. To address these needs, concentric tube robots have been investigated.  A robot design with up to four concentric tube arms, using three pre-curved tubes per arm, has been evaluated for reachability in a cadaver and for phantom tumor resection in anatomical skull models \cite{Weaver2012,Burgner2014}. The concentric tube robotic arms are particularly advantageous for this type of surgery because of their small size ($\diameter 2.32$ mm outer diameter) and tentacle-like dexterity. Furthermore, the addition of axial roll at the robot's wrist reduced the average resection time to 12.5 minutes and increased average removal percentage to 79.8 \% \cite{Swaney2015a}.
\par Another surgical target using transnasal access in the airways. This new surgical approach is motivated by the dichotomy of cost and complexity of the current trans-oral approach (which requires laryngeal suspension and full anesthesia with their associated post-operative sequelae) and the simplicity of micro-surgical tasks such as removal of polyps, cysts and nodules of the vocal folds or injection medialization as a treatment of unilateral vocal fold medialization.  Building on the concept of transnasal access into the airways as first proposed by Ikuta et al. \cite{ikuta2003_narrow_field_transnasal}, a rapidly deployable robot for transnasal micro-surgery was developed by Bajo et al. \cite{Bajo2013}. This robot uses a multi-backbone continuum robot architecture and it has a diameter of 5 mm with three working channels of $\diameter 1.8$ mm.  The unique aspect of this robot is its ability to actively comply with the environment when inserted through a  nasopharyngeal tube. This approach for active compliance of continuum robots was first proposed and developed by Goldman et al.\cite{goldman2014compliant}. This allows the surgeon to focus on the endoscope monitor when advancing the robot while not having to stop to steer the robot via a telemanipulation interface. Once in place, the robot can be telemanipulated to control an innection/aspiration needle, a biopsy cup and a flexible fiberscope.  The feasibility of using robot for transnasal surgery of the larynx and airways has been studied in mannequin and cadaveric models \cite{Dharamsi2014} and the safety of the cooperative insertion of the continuum robot was evaluated against a flexible fiberoptic laryngeoscope in Groom et al.\cite{Groom2015}. The results showed that, even though the multi-backbone robot is significantly stiffer than a flexible endoscope, when used for active compliance control it required insertion forces that were safe and close in value to a thinner and more flexible fiberoptic endoscope. 
\subsubsection{Transoral Systems}
\par Transoral surgery is challenging to perform due to the constrained workspace of the upper airways and the lack of distal dexterity provided by conventional tools. Robotic devices that improve distal dexterity could improve results in these procedures. In one study of treatments for orpharyngeal cancer, transoral robotic surgery had reduced morbidity and treatment costs while providing equivalent oncologic results when compared to conventional treatments \cite{dias2017role}. Several different robotic architectures have been presented for transoral surgery\cite{Simaan2004a, wang2008conceptual, Rivera-Serrano2012,Lang2017,he2013new,Olds2014}.
\begin{figure}[htbp]
    \centering
    \centerline{\includegraphics[width=1.0\columnwidth]{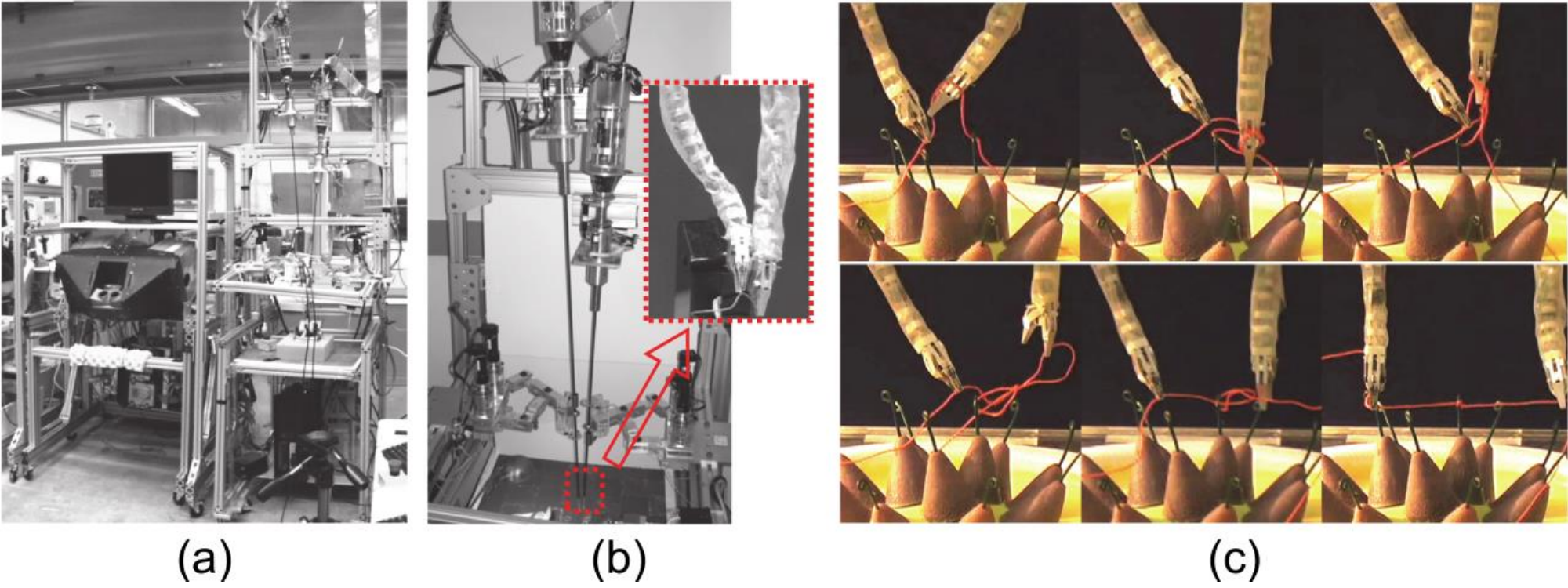}}
    \caption{A dual-arm system for trans-oral MIS of the upper airways: (a) Master-slave telemanipulation system, (b) slave arms, (c) dual-arm knot tying \cite{Hillel2008}}
    \label{Figure:transoral_simaan_taylor}
\end{figure} 
\par Multi-backbone continuum snake robots have been investigated for transoral surgery of the throat and upper airways \cite{Simaan2004a,Hillel2008, Simaan2009}. Such a system is shown in Figure \ref{Figure:transoral_simaan_taylor}-(a, b), and consists of bi-manual 4.2 mm diameter multi-backbone continuum arms with two stacked 2-DoF continuum segments per arm and a detachable gripper. Rotation of the continuum arms along the longitudinal axis, combined with synchronized actuation of the continuum segments, allows the end-effector to roll without changing the arm's configuration. Translation of the continuum arms provides an additional DoF, giving each arm 6-DoF. Actuation compensation was used to overcome modeling uncertainties, and the system was validated in a knot-tying experiment \cite{Simaan2009} as shown in \ref{Figure:transoral_simaan_taylor}-(c). Experiments reported by Hillel et al.\cite{Hillel2008} demonstrated the feasibility of Intracorporeal knot tying within the confines of a volume of an adult airways but the experiments were severely hampered by lack of depth perception. Triangulation of both arms to the same site was shown to be problematic due to the lack of a pitch/yaw wrist. Even though the continuum robots could provide this motion, the narrow airways meant that they would contact the sides of the airways when they bend to form "elbows" to allow for triangulation of both surgical arms to the same site.
\par
\begin{figure}[htbp]
    \centering
    \centerline{\includegraphics[width=0.8\columnwidth]{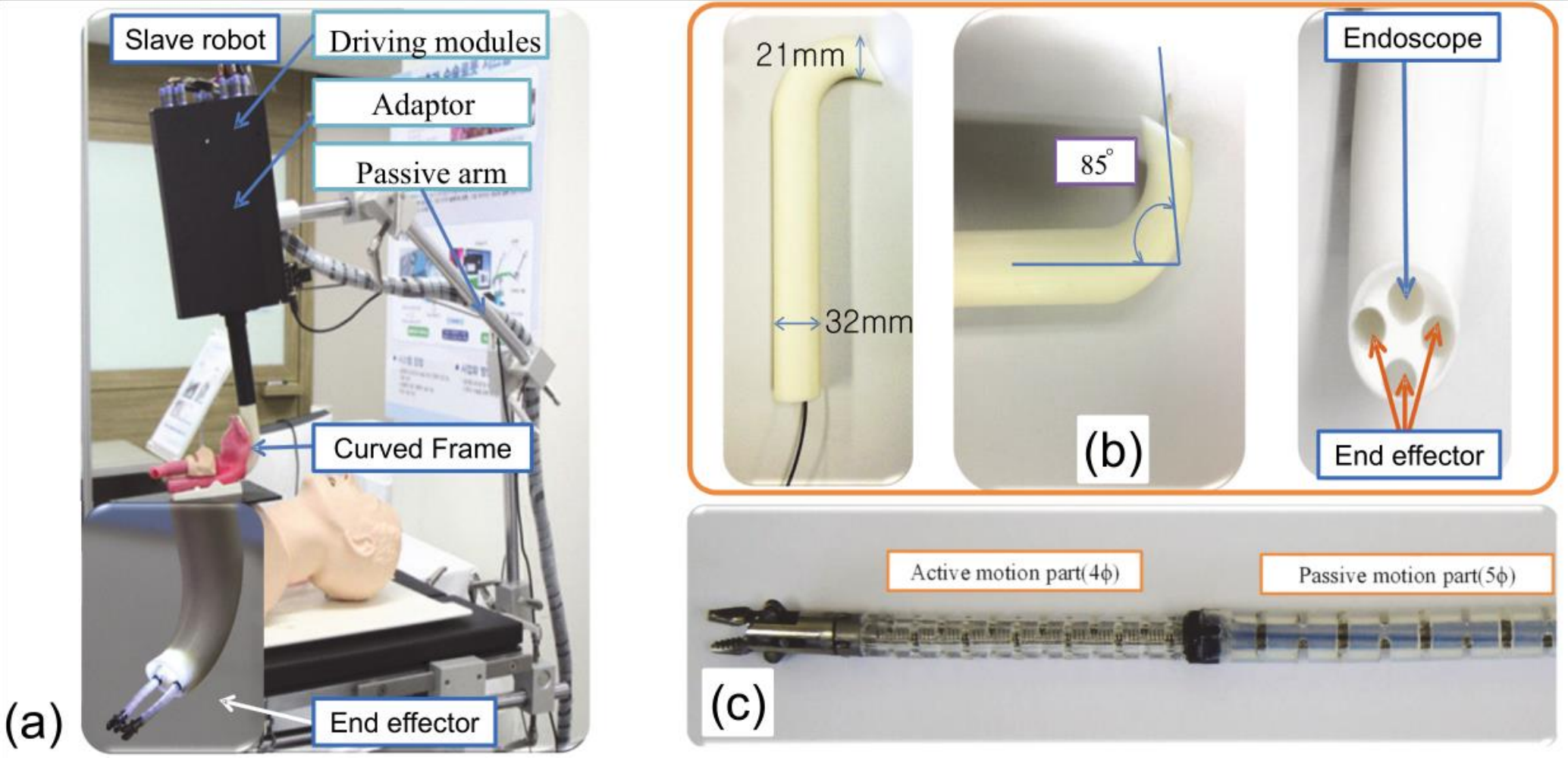}}
    \caption{(a) A slave robot for transoral surgery of the upper airways with a access overtube that provides the role of laryngeal suspension and provides access for robotic instruments, (b) a close-up view of a laryngeal suspension guide for inserting the continuum arms, (c) a two segment continuum arm. \cite{kwon2014suspension_TORS}}
    \label{Figure:bj_transoral}
\end{figure}

 Figure~\ref{Figure:bj_transoral} shows a proposed robotic device for suspension larygnscopy. This system consists of a passive positioning arm, a curved guide frame with two tool channels that provides a path to the larynx, and two flexible surgical instruments that pass through the guide frame channels \cite{kwon2014suspension_TORS}. The surgical instruments have a 5 mm diameter passive proximal section, which is constrained by the curved channel in the guide frame, and a 4 mm diameter robotically actuated distal section that exits from the guide tube and is used to perform the procedure. This system was used to successfully remove a polyp in a phantom model of the airways.
\par
\begin{figure}[htbp]
    \centering
    \centerline{\includegraphics[width=0.95\columnwidth]{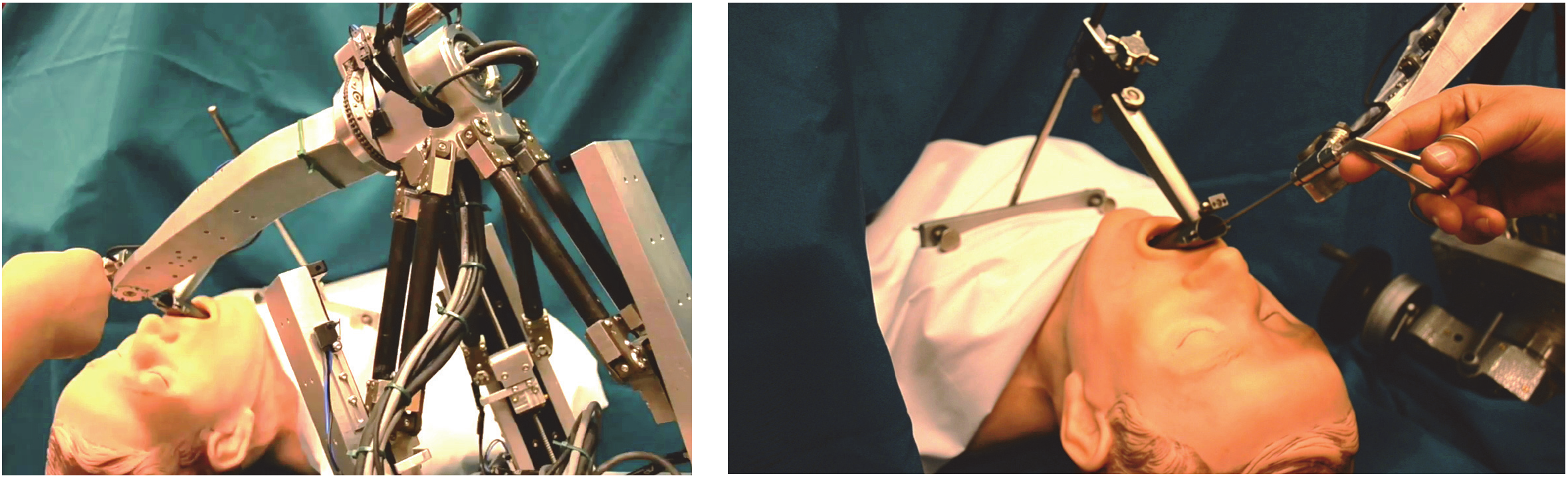}}
    \caption{The REMS cooperative manipulation robot for microsurgery of the upper airways\cite{Olds2014}.}
    \label{Figure:transoral_taylor}
\end{figure}

 A unique semi-active robotic system for transoral laryngeal surgery, shown in Figure \ref{Figure:transoral_taylor}, provides stabilization and assistive behaviors such as virtual fixtures and tremor canceling to improve surgical accuracy and precision, a concept that has been presented in earlier works\cite{taylor1999steady,jakopec2003_acrobot_cooperative_virtual_fixtures_davies2003}. This hybrid robot has a 3-DoF parallel delta robot that sits on a passive support stand in series with a robotic arm that holds the surgical instruments and provides two active rotary joints with passive roll of the tool. The use of the parallel robot for translations of the robot base results in increased stiffness and precision while the serial robot improves the reach of the robot, while still giving the surgeon control of the instrument actions \cite{he2013new,Olds2014}.
\par
\begin{figure}[htbp]
    \centering
    \centerline{\includegraphics[width=1.0\columnwidth]{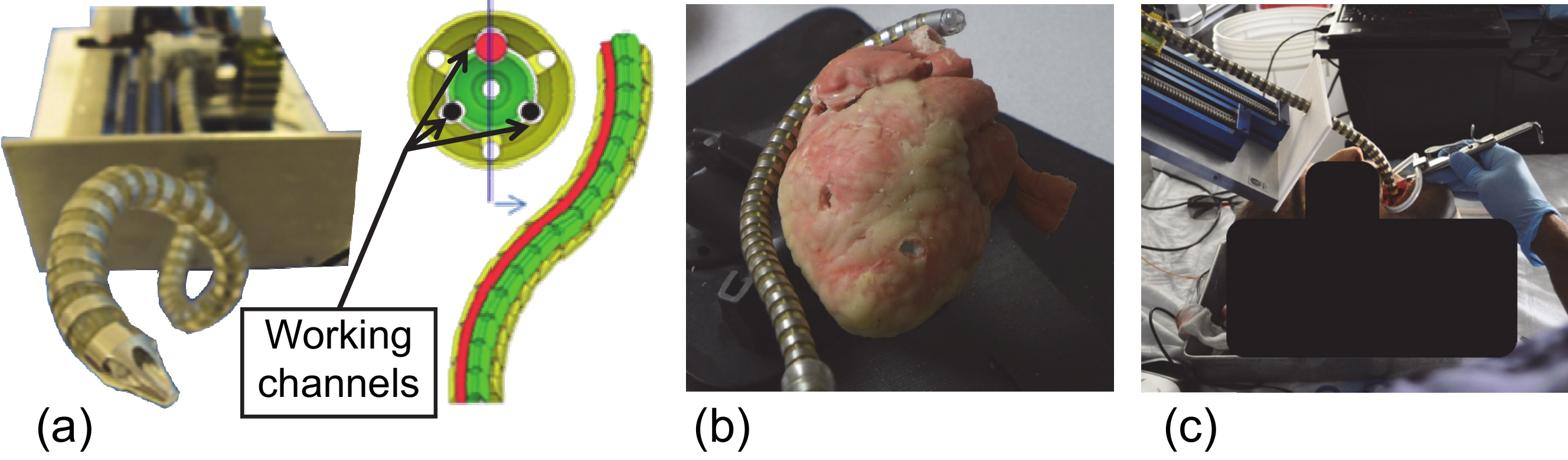}}
    \caption{The HARP robot: (a) The robots showing three working channels, (b) steering around a heart, (c) performing trans-oral access into the airways }
    \label{Figure:HARP}
\end{figure}

 The highly-articulated robotic probe (HARP), shown in Figure~\ref{Figure:HARP}, is a NOTES robot designed to provide deep access into the anatomy using cylindrical links for hemispherical ends that are steered and shape-locked using tendons in a manner generalizing the design concept presented in \cite{sturges1993flexible}. The device consists of an inner snake and an outer snake, which are able to achieve follow-the-leader motion by alternating steering and stiffening of each of the two snakes. This system was initially presented for epicardial ablation through a subxiphoid incision \cite{Ota2008}, but has recently been applied to transoral visualization and surgery of the upper aerodigestive tract \cite{Rivera-Serrano2012,Lang2017}. The device shown in Figure~\ref{Figure:HARP} provides three working channels for deploying ablation probes or flexible manual instrumentation. This technology is has been commercialized by MedRobotics Corporation as the Flex\textsuperscript{\textregistered} Robotic System and has recently received FDA clearance for transoral procedures in the mouth and throat. This system augments the concept of the HARP robot with manually actuated dexterous tools for intervention in the upper airways. Two manual instruments are positioned on the outside of a flexible guide robot resembling the HARP robot.  These instruments allow transmission of rotation along their backbone but lack a distal wrist with pitch and yaw capabilities.
\par Going farther down the airways, Sawaney et al.\cite{Swaney2015} proposed a system combining a wire-driven flexible bronchoscope, a concentric tube robot, and a steerable needle has been proposed for transoral lung biopsy. The bronchoscope is deployed transorally (either robotically or manually), a concentric tube robot is deployed through one of the bronchoscope channels and navigates past the bronchial wall, then a steerable needle is deployed through the concentric tube robot and is steered to the desired target. A coaxial tube can then be passed over the steerable needle to provide an access port for delivering therapies or performing a biopsy. This concept has been tested in a phantom model, and a motion planner\cite{Kuntz2015} has been introduced.
\par Another target application using trans-oral access is the surgical treatment of esophageal and gastric conditions and as a means for natural access for future trans-gastric abdominal surgery. There have been some developments of natural orifice translational systems including the following works. Abbott et al. \cite{Abbott2007} presented a second generation design of the ViaCath system,\cite{Abbott2007} which built on the Laprotek system by endoVia Medical Inc.. This system used a master interface with clutch, hold, and haptic feedback capabilities. Computer controlled instruments were carried to the surgical site by channels in a 3-DoF manually actuated steerable endoscope. The device facilitated the deployment of two surgical arms having a diameter of 7.2 mm. Each arm contained two nylon flexion points at the distal end with 2-DoF per flexion. A 1-DoF wrist and independent rotation of the tool jaws provides a total of 9-DoF to each instrument.  Actuation wire conduits are made from flexible close-wound spring shaft, and are twisted in opposing directions (and additional concentric conduits added) for higher torsional rigidity. Preliminary experiments with this system required the design of a $\diameter 19$ mm  over-tube for safe deployment of the gastroscope with the surgical arms.
\par Another system suitable for esophageal surgery and gastric surgery is the i-Snake \cite{Seneci2014}. This system has been developed as a highly articulated intraluminal surgery platform allowing safe access into the surgical site, visualization and deployment channels for surgical instruments. The system combines a wire-driven 2-DoF flexible proximal section and a 3-DoF articulated distal section with embedded motors. The most distal joint of the robot is a universal joint. The total diameter is 13 mm and the robot includes a camera, an optical fiber light, and a 3 mm working channel. A serial KUKA robot arm positions the snake robot relative to the patient and provides translation and rotation of the snake robot.
\par A prototype design of a NOTES  surgery system were presented by Min Seow et al.\cite{MinSeow2013} and Shen et al.\cite{Shen2015}. A similar friction shape-lock principle to the concept presented by Sturges\cite{sturges1993flexible} has been proposed for deploying miniature robotic modules with embedded actuation through the esophagus \cite{MinSeow2013}. A passively bending articulated snake-like arm having a diameter of 13 mm is used to support a multi-functional manipulator capable of introducing several tools when the snake-like arm is locked.  This design concept was later generalized by Shen et al.\cite{Shen2015} who replaced the passive bending articulated arm with a $\diameter 16$ mm wire-actuated articulated snake-like arm.  Both systems lacked distal dexterity wrists limiting their ability to effectively carry out dexterous dual arm tissue manipulation tasks such as suturing or knot tying.
%
\subsubsection{Transurethral Systems}
\par The transurethral access route is suitable for surgical intervention in the prostate, bladder, ureters and kidneys. Robotic intervention has been generally limited to prostate cancer resection and bladder cancer resection.
\par Current techniques utilizing manual tools suffer from insufficient resection accuracy due to the lack of tooltip dexterity and reduced access to the anterior bladder wall. Furthermore the limited tool accuracy in controlling depth of tumor resection, the limited depth perception and the conscious decision to minimize risk of perforating the bladder walls cause surgeons to generally under-resect tumors \cite{sarli2016_resection_bladder_ijmrcas}. These factors, along with difficulty in discerning the extent of non-muscle invasive tumors and lack of dexterity allowing tumor resection \textit{en block}, are considered factors in the high  recurrence rate of bladder cancer. To overcome some of these challenges robotic systems have been developed to improve dexterity and reach of the surgical instruments, which could improve resection accuracy, reducing the need for re-resection \cite{Herrell2014}.
\begin{figure}[htbp]
    \centering
    \centerline{\includegraphics[width=0.95\columnwidth]{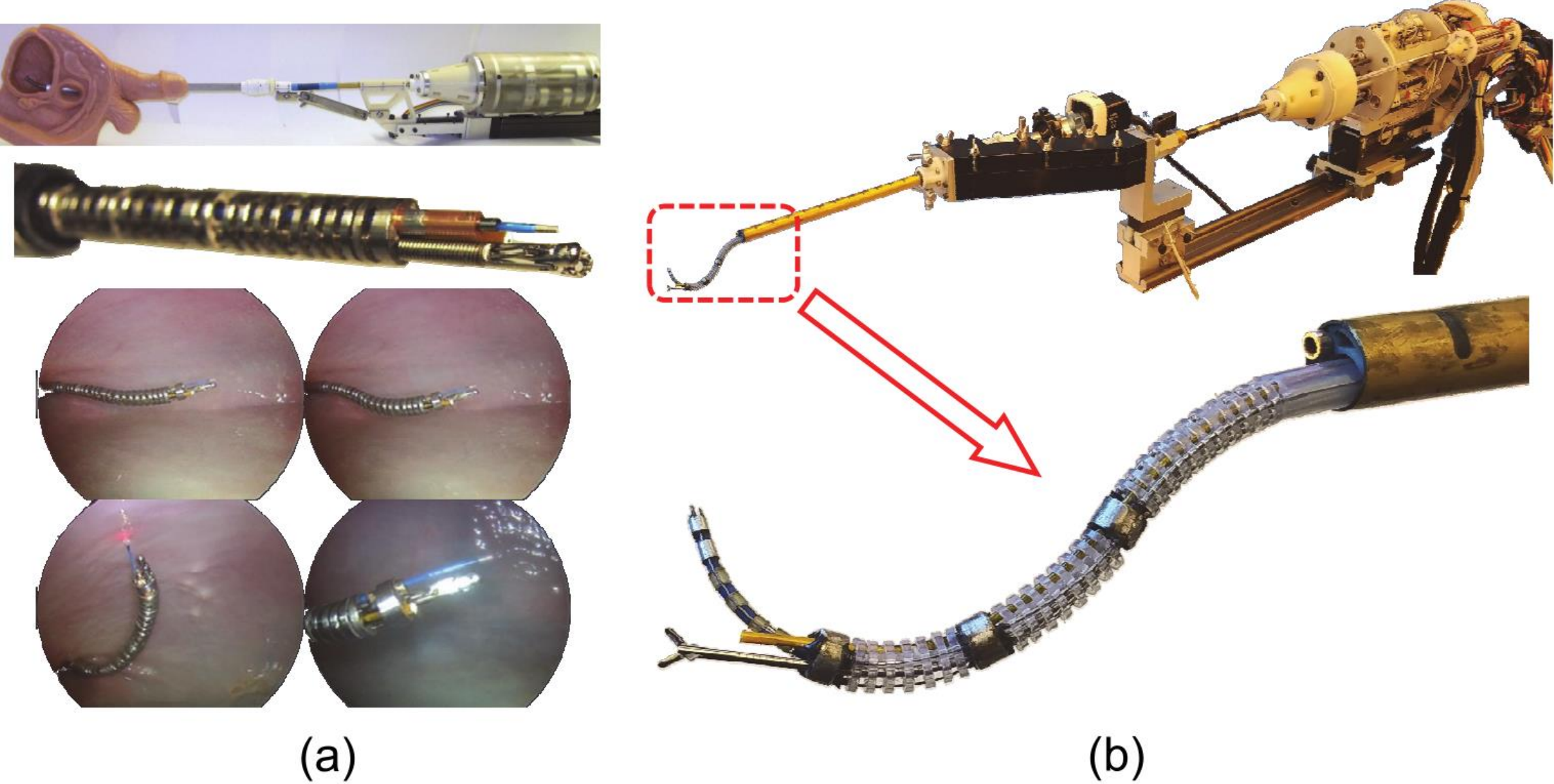}}
    \caption{A transurethral bladder cancer resection system (a) first prototype without independent laser control\cite{Goldman2013}, (b) second prototype with a custom resectoscope\cite{Sarli2016}, a fixed endoscope and a 1.6 mm continuum snake for controlling the laser. Both prototypes allow deployment of a fiberscope, a biopsy cup/graspers and an ablation laser fiber}
    \label{Figure:transurethral_simaan}
\end{figure} 
\par A robotic system for TURBT, shown in Figure \ref{Figure:transurethral_simaan}-(a), consists of two stacked continuum segments, each having three secondary NiTi push-pull backbones and an outer diameter of 5 mm \cite{Goldman2013}. 2-DoF per segment plus insertion of the manipulator provides a total of 5-DoF. Lumen within the continuum segments carry a fiberscope, a laser cautery fiber, and biopsy forceps for tumor resection. A 7-DoF actuation unit with DC motors and lead screws allows for insertion of the resectoscope and actuation of the continuum segments. The system can achieve sub-millimeter positioning accuracy, and provides force sensing capabilities using load cells at the base of each secondary backbone within the actuation unit. An \textit{ex vivo} study with bovine bladder showed this system was able to accomplish surgical tasks during TURBT, including visualization of the bladder walls, energy delivery, biopsy, and tumor resection. An upgraded design of this robot, shown in Figure \ref{Figure:transurethral_simaan}-(b), includes independent laser control, three stacked continuum segments, a polytetrafluoroethylene elastomer backbone, a custom resectoscope, and a modular actuation unit that partially decouples the actuation of the secondary backbones \cite{DelGuidice2016}.
\par Benign prostatic hyperplasia (BPH) is a nonmalignant enlargement of the prostate that results in restriction of urine flow through the urethra and in severe cases can lead to complete blockage and kidney damage. Transurethral resection of the prostate (TURP) is the current minimally-invasive standard of care for BPH \cite{marszalek2009transurethral}, with holmium laser enucleation of the prostate (HoLEP) a potentially promising alternative to TURP \cite{mandeville2011new}. The lack of distal dexterity while operating on the prostate makes these procedures technically challenging \cite{Hendrick2014}. This technical difficulty manifesting in limited accuracy of resection and difficulty in enucleating tissue with minimal tilting of the rigid tools and the urethral anatomy has motivated the investigation of using robotic assistance for transurethral resection of the prostate.
\begin{figure}[htbp]
\center
\includegraphics[width=0.7\textwidth]{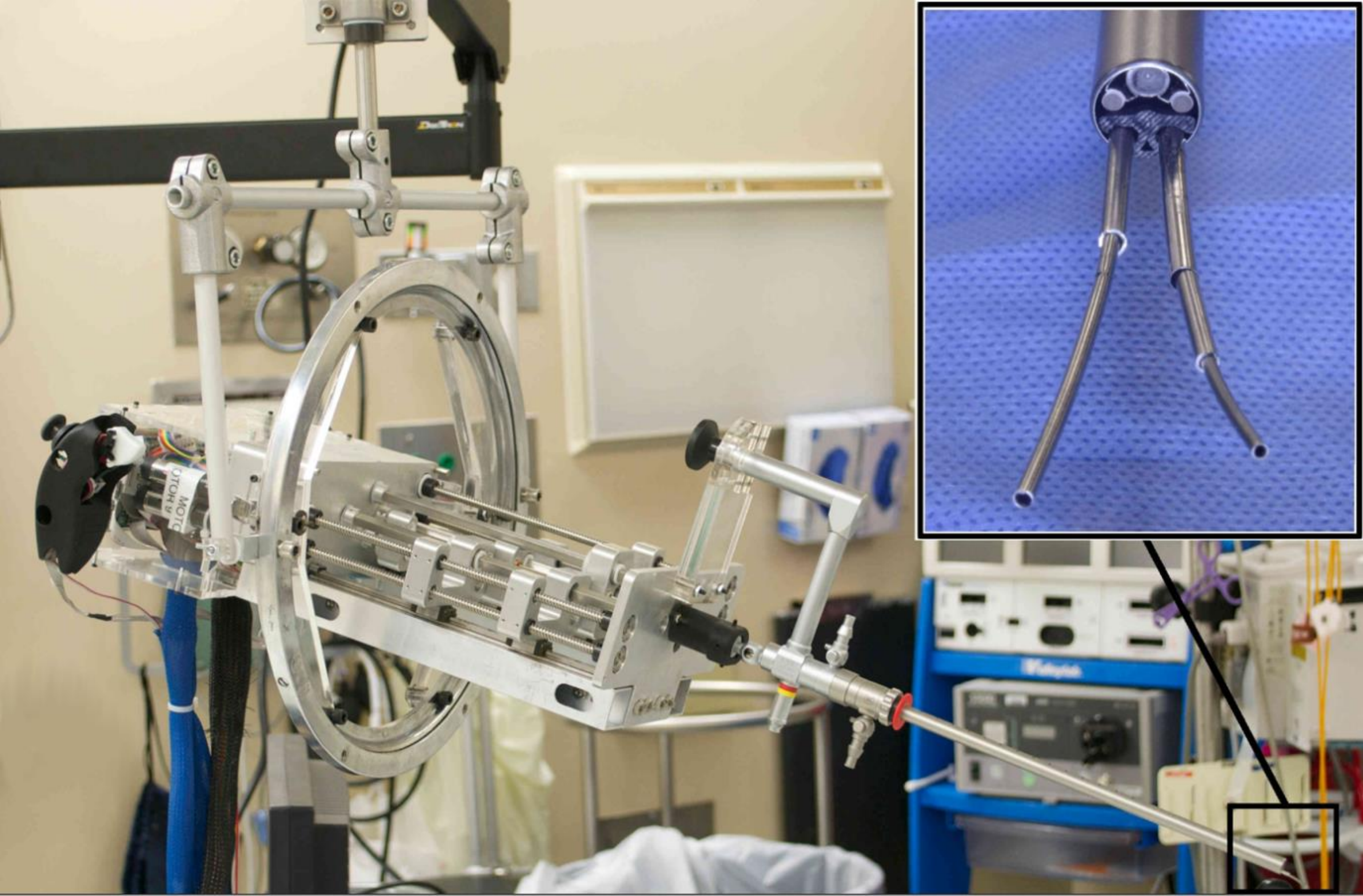}
\caption{This robot for transurethral HoLEP consists of a hand-held user interface and concentric tube manipulators that are actuated by an electormechanical actuation unit. Reproduced with permission \cite{Hendrick2015}.}
\label{fig:webster_system}
\end{figure} 
\par Transurethral laser surgery of the prostate was first carried out by Davies et al. \cite{davies1989_first_transurethral_prostate} in 1988 and followed with series of works using an eight degree of freedom robot comprised of a Puma 560 robot\cite{Davies1991, davies1993prostate_transurethral} that was augmented with two more degrees of freedom. This seminal work demonstrated the utility of robotic assistance for achieving accurate resection, but did not solve the problem of distal dexterity.  Recently,  concentric tube manipulators have been applied to transurethral prostate procedures. A concentric tube system\cite{Hendrick2014} for holmium laser enucleation of the prostate (HoLEP), a laser-based alternative to transurethral resection of the prostate (TURP), is shown in Fig. \ref{fig:webster_system}. It consists of a rigid endoscope with two concentric tube manipulators that are passed through the endoscope's tool channels. One manipulator has 6-DoF with three pre-curved tubes, while the second has 3-DoF with two pre-curved tubes. A hand-held user interface allows the surgeon to manually control the position of the endoscope, and thumb joysticks and triggers control the position of each concentric tube manipulator in either joint space or task space. Initial cadaveric experiments showed this system was capable of executing tasks necessary for HoLEP \cite{Mitchell2016}.
\par Russo et al.\cite{Russo2015} presented another robot for laser-assisted transurethral resection of prostate, called the ASTRO (actuated and sensorized tool for laser-assisted surgery of the prostate). This robot uses a wire-driven multi-lumen catheter to steer a laser-fiber and fiber-optic based contact sensing. This system has thus far been tested \textit{in vitro}.
%
\subsubsection{Transanal Systems}
\par Although still a new application for robotic surgery, transanal minimally invasive surgery has been shown be a safe and feasible application for use with the da Vinci\textsuperscript{\textregistered} Si\cite{Albert2015}. Further improvement in distal dexterity using snake-like devices has the potential to allow for more complex procedures to be done and for sites further within the colon to be accessed.
\par The distal highly articulated sections of the bi-manual micro-IGES system \cite{Shang2017} provide increased dexterity for transanal micro-surgery. Although the shafts are straight and rigid, the wire-driven articulated section for this system's instruments can have up to 7-DoF. The complete system, including two haptic input devices, has been tested in bovine bowel tissue for marking and suturing lesions, and for complete transanal endoscopic micro-surgery in live porcine models.
\par Another robotic system for transanal surgery has two manipulators that exit from a conventional endoscope in a triangulated configuration \cite{Phee2009}. One manipulator has a cauterizing hook end effector, and the other has a gripper. Both consist of a wire-driven linkage and provide a total of 9-DoF. The total diameter of the robot is 16 mm. The system was been tested \textit{in-vivo} for endoscopic sub-mucosal dissection (ESD), transgastric liver wedge resection, and full-thickness gastric resection \cite{Chiu2014,Chiu2015}. This technology is currently being commercialized by EndoMaster Pte. Ltd.
\par Due to this need for distal dexterity, parallel continuum robots\cite{bryson2014toward} have been proposed for use in transanal endocolonic surgery. These robots can provide 6-DoF dexterous manipulation in a single continuum segment, making them a potential candidate for the narrow, constrained workspace of the colon. A 12 mm diameter parallel continuum manipulator with a 2-DoF grasper has been presented \cite{orekhov2016analysis}.  These new design concepts remain to be validated and tested in clinical application scenarios.  

\section{Future Trends}
\par
\begin{figure}[htbp]
    \centering
    \centerline{\includegraphics[width=0.4\columnwidth]{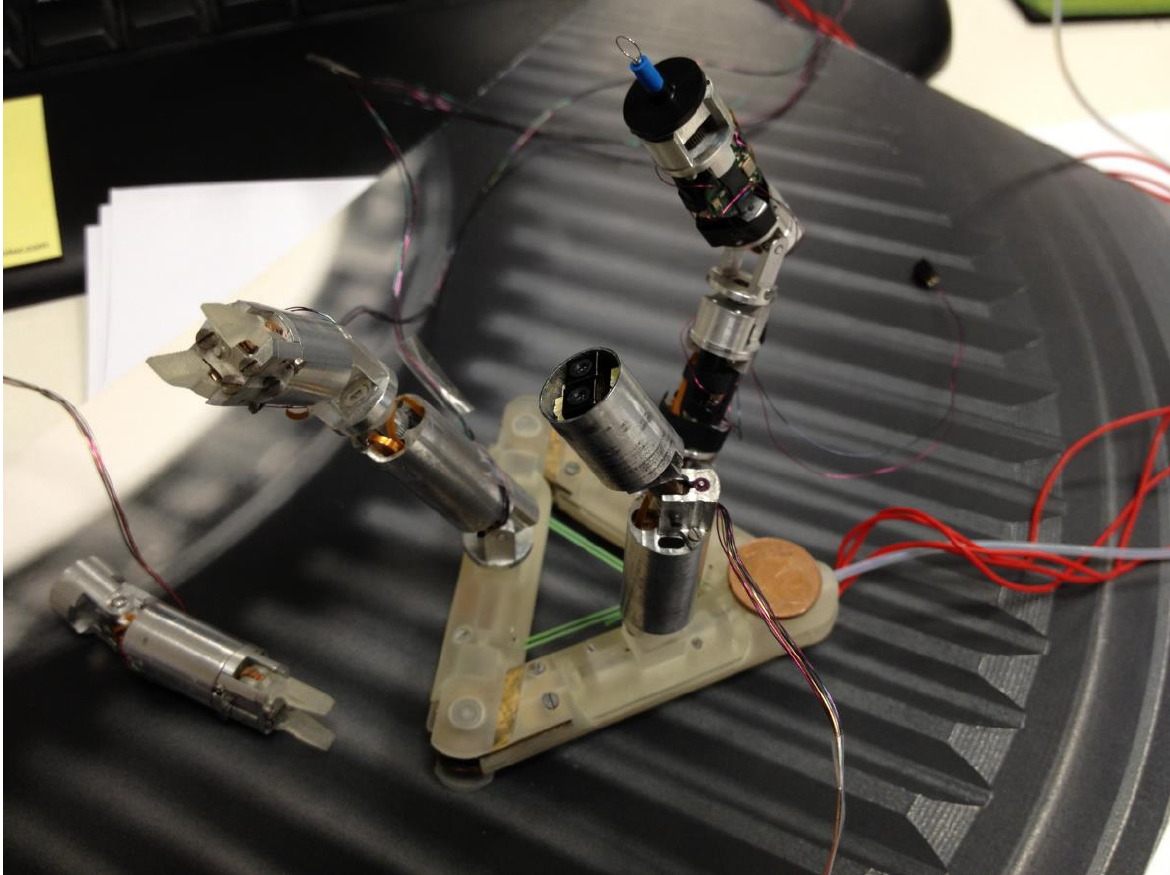}}
    \caption{A magnetically-anchored assemblable robot showing a magnetically anchored base with three working arms\cite{Tortora2014}}
    \label{Figure:magnetic_notes_arianna}
\end{figure}

 In addition to the works considering snake-like robots for surgery, there have been recent advancements in two key technologies that hold promise: soft robotics and modular magnetically actuated robots. Several works have explored the design space for reconfigurable robotic systems that are assembled after being deployed into the surgical workspace \cite{Lehman2009,Tortora2014,Tognarelli2015}. The system shown in Figure \ref{Figure:magnetic_notes_arianna} has 12 mm diameter robotic modules, each of which may have either manipulation, cutting, vision, or retraction functions, deployed through the esophagus and mounted on a frame that is anchored to the abdominal wall with external permanent magnets. After insertion, the modules are assembled to the anchoring frame using an assistive endoscope. The anchoring frame, which is 14 mm in diameter, is inserted in a straight configuration and transitions to the triangular configuration using shape-memory alloy springs \cite{Salerno2013}. Benefits of this robotic architecture is the removal of the kinematic constraint created by having to operate through an incision port or natural orifice, increased triangulation, and a potentially improved ability to access hard-to-reach anatomy.
\begin{figure}[htbp]
    \centering
    \centerline{\includegraphics[width=0.95\columnwidth]{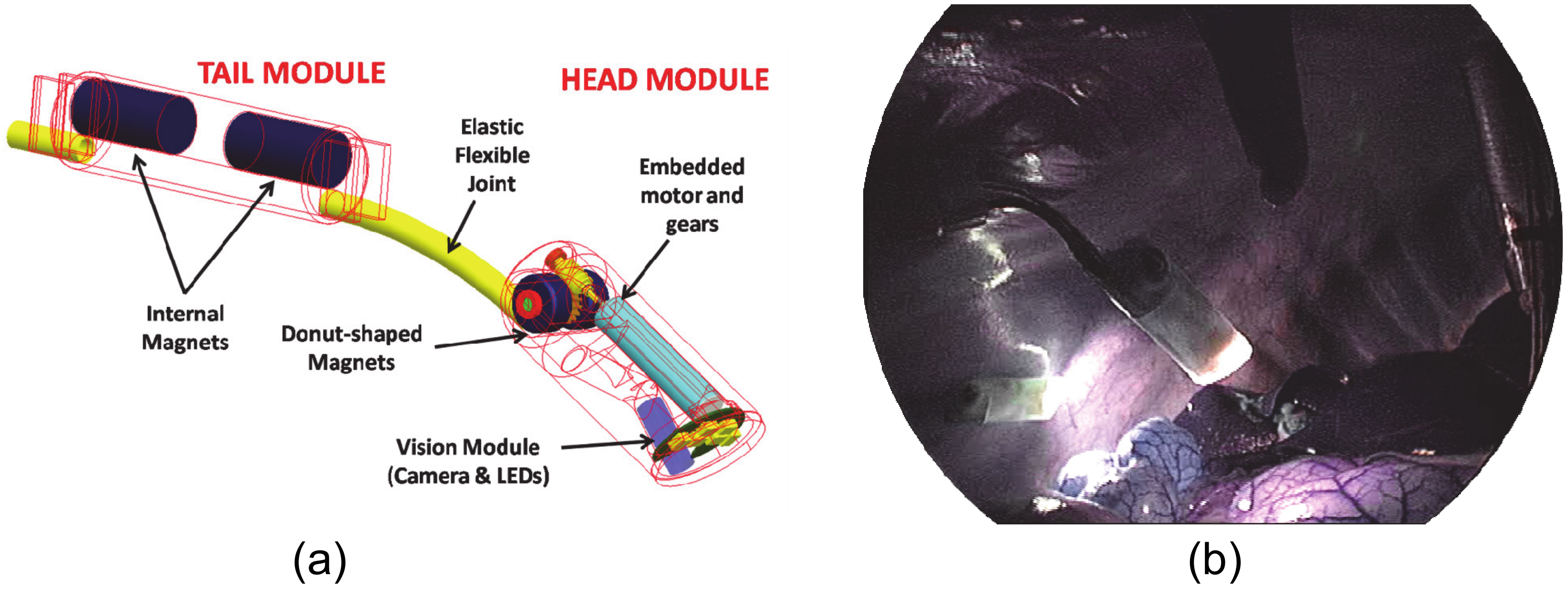}}
    \caption{(a) Concept for a magnetically actuated compliant joint (b) In-vivo deployment of the magnetically actuated surgical robot \cite{DiNatali2015}}.
    \label{Figure:magnetic_actuation}
\end{figure}
%
\par In addition to magnetic anchoring of miniature robots, there are a number of works exploring the use of magnets for actuation, as reviewed in \cite{Leong2016}. After anchoring a tool magnetically, an additional external magnet can apply forces to internal magnets to provide torque or actuate mechanisms that control surgical tools. One work showed that this scheme could provide up to 13.5 mNm of torque in closed-loop actuation \cite{DiNatali2015}. A concept for a compliant joint, shown in Figure \ref{Figure:magnetic_actuation}, used fixed external magnets, an internal magnetically anchored fixed end, and a free end with an embedded motor that actively rotated internal magnets to actuate the free end. This magnetic actuation scheme was used to orient an endoluminal camera \cite{Simi2013}. Magnetic actuation could be useful for both diagnostic purposes and for performing surgical tasks, as shown by work where a tethered capsule with a camera, permanent magnet, magnetic field sensor, and two tool channels was used to navigate and inspect a porcine colon before performing resection \cite{Valdastri2012}.
\par Another new design space for surgical robots is variable stiffness soft robots with pneumatic actuation. Single-port and NOTES surgical robots require sufficient stiffness to manipulate tissue, but it is also beneficial for them to be compliant to improve their inherent safety and reduce trauma during insertion. One way to achieve this is by incorporating mechanisms that vary the manipulator's stiffness, allowing for both compliant insertion of tools and stiff manipulation of tissue. The STIFF-FLOP manipulator \cite{Cianchetti2013} uses the mechanics of granular jamming to modulate stiffness in a soft, silicone manipulator. Each soft module contains three pneumatic chambers arranged lengthwise along the module, which when pressurized, induce a curvature. The module is furthermore covered with a braided sheath to prevent outward expansion of the pneumatic chambers. A central chamber is filled with a granular material that interlocks when a vacuum is applied, increasing the stiffness of the manipulator. Recently, another stiffening mechanism applied to these soft robots used tendons actuated antagonistically to the pneumatic actuation to achieve up to 94\% increase in stiffness, compared to a maximum 41\% increase for the granular jamming robot \cite{Shiva2016}. Dynamic and stiffness models for these robots have also been presented. \cite{Renda2014,Sadati2016}

\section{Open Problems in Robotics for MIS, SPA and NOTES}
\par Minimally invasive surgery in confined spaces has advanced significantly in the last 15 years. Thanks to prior exploratory works in robotics for intraluminal and extraliminal minimally invasive surgery, we have gained significant understanding of test requirements, design specifications and design concepts capable of meeting the minimum requirement of surgery, i.e. workspace and distal dexterity. Despite the significant progress made, there are still key technical hurdles to successful adoption and deployment of robotic systems in less invasive intraluminal and single port access surgeries. These challenges are hereby summarized with suggestions for future areas of research.

\par The first unmet challenge is still the lack of formal design methodology of robots for MIS, SPA and NOTES. Because engineers and surgeons are still at exploratory stages of these surgical paradigms, designers are often forced to rely on ad-hoc design decisions with trial and error. There is a significant lack of literature documenting anatomical manipulation requirements including forces required for tissue retraction, blunt dissection, and required minimal reachable workspace and dexterous workspace for a given target organ. More works on generating organ-specific tissue models and digital shared sets would greatly facilitate robotics research in the formation of a formal design methodology. The Visible Human Project by the National Library of Medicine is a good start in terms of curating a 3D model of a male and female adult anatomy, but it is still not detailed enough or formed in a model that can be easily annotated with tissue characteristics including stiffness.
\par The second part preventing the successful deployment of new surgical systems methodology is the fact that new surgical systems for SPA, NOTES and intraluminal surgery generally push the boundaries of the traditional design space. The use of wire-actuated snake-like and continuum robots has facilitated excellent first stage proof of concept systems, but our current understanding in modeling frictional and actuation losses in these systems is still a hurdle. Simple design questions such as determination of dynamic motion bandwidth by some of these new robot architectures will often result in very hard problem formulations in terms of dynamics, mechanics and control modeling and design. Alternative approaches for use of sensory information for mixed feedback control\cite{bajo2011configuration} or the recently proposed model-less control framework\cite{yip2014model} are promising new approaches that remain to be further explored. More research in the area of control and modeling of these robots is essential for obtaining high quality motion and force control of these robots.
\par Besides robot design, there still are fundamental challenges in terms of human-robot interaction, sensing and high-level telemanipulation control. These challenges include the fact that we currently do not have a good framework for designing telemanipulation master devices and user interfaces that are suitable for the highly articulated and branched/multi-arm robot architectures needed to address the requirements of NOTES, SPA and intraluminal surgery. This necessitates the rethinking of how high-level telemanipulation interfaces should be used to help surgeons achieve the surgical outcomes for their patients. Even though there have been works on cooperative manipulation of surgical instruments using semi-active robots such as the Steady-Hand robot\cite{taylor1999steady} or the Acrobot\cite{jakopec2003_acrobot_cooperative_virtual_fixtures_davies2003}, these concepts of human-robot interaction using assistive control laws (known as virtual fixtures\cite{rosenberg1993virtual}) are hard to translate within the context of SPA, NOTES and intraluminal surgery. The difficulty arises from the fact that surgical robots for these new surgical paradigms have to interact with the anatomy along their entire length. As a result, there is a new need for expanding the framework for defining these virtual fixtures to take into account constraints along the robot body. More importantly, there is a need to define new path planning and control and sensing strategies and technologies to allow the in-vivo characterization of the allowable motion space of these robots so that virtual fixtures can be defined in order to assist the user to safeguard the anatomy. Finally, just as in open surgery where multiple surgeons can collaborate of a surgical task, there is a need for new telemanipulation frameworks allowing effective collaboration of at least two surgeons - despite the fact that they have sensory and perception deficiencies regarding the nature of the robot interaction with anatomy in points that are outside the visual field of these robots.
\pagebreak
\par At the level of sensory acquisition and feedback, there have been many works (e.g. \cite{perri2010initial_localization_palpation, tholey2005force_feedback_desai, wagner2002force, kitagawa2002analysis}) demonstrating the importance and value of force feedback to the users. Some of these works were partly inconclusive due to the quality of the haptic feedback\cite{okamura2009haptic}, but it is clear that having high quality force feedback is useful and helpful to surgeons to achieve consistent forces of interaction with the tissue and for uniform knot tying, to safeguard against accidental trauma and to help localize tumors. The vast majority of surgical systems today still do not have force feedback. While this issue is not critical in multi-port MIS, it is highly important in NOTES, SPA and intraluminal surgery where the visual perception barriers are even stricter. There have been some recent results in obtaining indirect estimation of forces on continuum and surgical robots\cite{Xu2008, puangmali2008state, haraguchi2011prototype, rucker2011deflection, wei2012modeling} or using direct sensing via miniature force sensors\cite{seibold2005prototype, baki2013design}. However, these exploratory solutions have not made it into clinical practice either due to cost and sterilization limitations of dedicated sensory technology or because of complexity of the indirect force estimation algorithms and the uncertainty encumbered in modeling and accounting for friction.
\par Finally, many of the surgical paradigms still fail due to lack of perception and in-vivo sensory information allowing the surgeons to correlate the surgical scene with pre-operative plans. To overcome the fact that organs shift and swell during surgery, recent works such as \cite{chalasani2016concurrent, wang_2017_JMR, ayvali2016_probing_icra, pile2014force} have started to explore the use of in-vivo model update based on adapting a pre-operative to a model created using sensory data including force, contact and stiffness. These approaches complement prior works on using geometric scanning and registration of organs (e.g. \cite{miga2003cortical, burgner2013study, hayashibe2006laser}). Despite progress made in these works, the problem of incorporating in-vivo sensory data to guide and improve the surgical process still stands. 
\section{Conclusion}
This chapter reviewed the recent works on minimally invasive surgery with specific emphasis on single port access and intraluminal surgeries. The design requirements and challenges presented by these new surgical paradigms have been presented. A brief exposition of snake-like robot architectures for meeting the manipulation demands of these new surgical paradigms has been outlined along with sample systems in each one of the sub-surgical domains of extraluminal single port and intraluminal surgery. The overview of the works highlights the diverse set of solution approaches and the significant progress made towards providing exploratory investigations of robotics in these new surgical paradigms. Challenges and knowledge gaps limiting successful clinical adoption of new surgical systems for single port access and intraluminal surgery have been presented. 

\section*{Acknowledgments}
A. Orekhov is funded by the National Science Foundation Graduate Research Fellowship Program under Grant No. 1445197. Any opinions, findings, and conclusions or recommendations expressed in this material are those of the authors and do not necessarily reflect the views of the National Science Foundation.
\bibliographystyle{ws-rv-van}
\bibliography{ms}
\end{document}